\documentclass[journal]{IEEEtran}

\usepackage{xcolor}
\usepackage{url}
\usepackage{hyperref}
\usepackage{amsmath,amsfonts}
\usepackage{graphicx}
\usepackage{enumitem}
\usepackage{soul,color}
\soulregister\ref7
\usepackage[flushleft]{threeparttable}
\usepackage{adjustbox}
\usepackage{booktabs}
\usepackage{multirow}
\usepackage[
  separate-uncertainty = true,
  multi-part-units = repeat
]{siunitx}

\usepackage[linesnumbered,ruled,vlined]{algorithm2e}
\usepackage{xcolor,soul} 

\definecolor{yellow_periko}{cmyk}{0.3, 0.1, 0.3, 0.001}

\sethlcolor{yellow_periko}

\def\HiLi{\leavevmode\rlap{\hbox to \hsize{\color{yellow_periko}\leaders\hrule height 0.7\baselineskip depth .5ex\hfill}}}

\begin{document}




\title{A Graph-based Methodology for the Sensorless Estimation of Road Traffic Profiles} 


\author{Eric L. Manibardo,
        Ibai La\~na,
        Esther Villar-Rodriguez,
        and Javier Del Ser,~\IEEEmembership{Senior Member,~IEEE}
        \thanks{Eric L. Manibardo, Ibai La\~na, Esther Villar Rodriguez and Javier Del Ser are with TECNALIA, Basque Research and Technology Alliance (BRTA), 48160 Derio, Bizkaia, Spain.
        Javier Del Ser is also with the University of the Basque Country (UPV/EHU) 48013 Bilbao, Bizkaia, Spain.}
        \thanks{e-mail: [eric.lopez, ibai.lana, esther.villar, javier.delser]@tecnalia.com}
        }

\maketitle

\begin{abstract}
Traffic forecasting models rely on data that needs to be sensed, processed, and stored. This requires the deployment and maintenance of traffic sensing infrastructure, often leading to unaffordable monetary costs. The lack of sensed locations can be complemented with synthetic data simulations that further lower the economical investment needed for traffic monitoring. One of the most common data generative approaches consists of producing real-like traffic patterns, according to data distributions from analogous roads. The process of detecting roads with similar traffic is the key point of these systems. However, without collecting data at the target location no flow metrics can be employed for this similarity-based search. We present a method to discover locations among those with available traffic data by inspecting topological features. These features are extracted from domain-specific knowledge as numerical representations (embeddings) to compare different locations and eventually find roads with analogous daily traffic profiles based on the similarity between embeddings. The performance of this novel selection system is examined and compared to simpler traffic estimation approaches. After finding a similar source of data, a generative method is used to synthesize traffic profiles. Depending on the resemblance of the traffic behavior at the sensed road, the generation method can be fed with data from one road only. Several generation approaches are analyzed in terms of the precision of the synthesized samples. Above all, this work intends to stimulate further research efforts towards enhancing the quality of synthetic traffic samples and thereby, reducing the need for sensing infrastructure.
\end{abstract}


\section{Introduction}
\label{sec:intro}

One of the main goals of traffic management is to provide a safe traffic infrastructure, where most of the potential traffic congestion scenarios and car collisions are avoided. This issue can be addressed by modeling the traffic behavior of the road network and by developing predictive methods that allow forecasting metrics that characterize traffic, such as flow, speed, or travel time \cite{lana2018road}. This process enables the implementation of operational measures for decision making (e.g., redirecting traffic flow to other alternative routes). These traffic forecasting models are usually built from past traffic observations. However, in practice traffic data acquisition systems cannot be deployed over every link of a road, mainly due to the high costs of deployment and maintenance of such sensing equipment. On many occasions this issue has been addressed by deploying temporal sensors that provide measurements for certain locations of interest during a limited period of time. Nevertheless, a proper characterization of the traffic behavior under a variety of circumstances (e.g., events or holidays) requires real traffic measurements over more dilated periods \cite{manibardo2020transfer}. Thus, real traffic data are not available for every road link, nor are they collected for the time needed for a thorough characterization. Indeed, in some cases placing a sensor is not feasible due to manifold reasons. If possible, another concern arises from the long latency needed to collect sufficient data for the subsequent model training. Since the performance of traffic forecasting models is constrained to both data quality and quantity, \emph{data availability} is arguably the practical bottleneck for a successful traffic network characterization.

For those scenarios where data are available, a glance at the state-of-the-art reveals that the community is close to the upper bound in terms of model-based improvements, whereas data quality is acknowledged to be the key for more accurate predictions \cite{manibardo2021deep}. In the context of traffic modeling, data quality \cite{pipino2002data} can be regarded in the form of anomaly absence. Some of the most widely observed anomalies include missing traffic records, or data aggregation failures that produce outliers shaped as spikes over the traffic time series.

These anomalies are strongly influenced by the sensing technique in use. On one hand, Roadside Car Data (RCD) is obtained by deploying sensors at particular locations of the traffic network, such as inductive loops or traffic cameras. These sensors continuously capture the traffic flow or mean speed at the point where they are deployed. On the other hand, Floating Car Data (FCD) refers to traffic measurements of individually tracked vehicles, via devices with positioning sensors (e.g., GPS). This type of data can provide more precise traffic measurements. However, the processing pipeline needed for modeling them effectively poses several challenges, such as i) the segmentation of distinct travel modes, as subjects often mix them over the same trip; ii) human-induced mistakes during the data annotation process; and iii) intrinsic precision errors of the positioning sensor. 

The deployment and maintenance costs of RCD, together with the concerns associated to FCD, restrict the number of public traffic datasets that ensure an acceptable level of data quality and enough time span to cover traffic profiles occurring along the year. As a result, the community often resorts to a reduced collection of datasets (with above-average data quality and time coverage) for research and industrial purposes \cite{manibardo2021deep}. Examples of these popular data repositories are the Caltrans Performance Measurement System (PeMS) data source \cite{PeMSddbb}, the Beijing Q-Traffic dataset \cite{liao2018deep}, or the Los Angeles County highway (METR-LA) data source \cite{jagadish2014big}. Unfortunately, building predictive models from a limited collection of datasets is a trend that leads to scarce insights beyond those that have already been given in the prior literature. When it comes to model benchmarks where the goal is to design a top-leading traffic predictive method (in the form of a new model architecture or data preprocessing technique \cite{lana2021data}), the selection of well-known datasets eases the reproducibility of the results, ensures the transparency of the reported performance scores, and stimulate follow-up studies departing therefrom. However, from a practical point of view there are no better traffic datasets than those collected over the same traffic network where the forecasting system is aimed to. Thus, generative methods and simulators can provide synthetic traffic datasets tailored for the same context as the target location.

In summary, the nature of RCD and FCD limits the disposal of traffic data at every point of a traffic network. In turn, simulators are hard to adjust, and without a precise configuration real-world factors that influence traffic can not be modeled. Nevertheless, synthetic data can be used by traffic managers to extract insights from locations of the traffic network where no sensors can be deployed, nor simulations can be performed. 

Partly inspired by the findings of \cite{lana2021soft}, in this work we present a novel method that allows to find a road segment with a similar traffic behavior in relation to a non-sensed road segment. In this context, no traffic data is available at the target road, so the only information that can be used to make the selection has to be extracted from the design of the traffic network and other circumstantial information that can be extracted from its context. Under the concept of \emph{road feature embedding}, we design a set of features that attempt to characterize each road segment, based on the topology and context of the area and the domain-specific knowledge about the behavior of traffic flows in similar urban areas to the ones under study. Our hypothesis is that locations with similar topological and contextual features (and thus, road feature embeddings) are more likely to share traffic patterns. It can be expected that a secondary road has a higher traffic flow than a residential avenue (and vice versa). The design of the traffic map itself determines the capacity of each road and influences how drivers behave when driving therethrough. However, the number of lanes and road type are not the unique traits that affect the traffic profile of a certain road segment: the location and connections to other road segments can also bias its traffic profile. Intuitively two avenues sharing the same road type, number of lanes and other topological characteristics (e.g., lane width) might display different traffic flow measurements depending on which other roads they are connected to. They can receive traffic from main arterial roads, but also residential areas. Two equally designed parallel roads can exhibit opposite traffic profiles, where one presents congestion during the morning commute to work, and so does the other road in the afternoon. Human intuition and experts in traffic management can produce estimates of the traffic behavior to a point where a person can produce its own congestion predictions towards selecting the best route during a city trip. In essence, the main goal of this work is to \emph{translate} this knowledge into a numeric vector (namely, a road feature embedding), so road segments of similar traffic profiles can be found without comparing traffic measurements, but the aforementioned expert knowledge-driven features instead.

Our analysis is supported by a collection of 55 Automatic Traffic Readers (ATR) 
deployed at the city of Madrid, Spain, providing traffic patterns of distinct shape and scale. However, while delving into the process of finding road segments with analogous daily traffic profiles, a research question arises: if the chosen dataset already has a similar traffic profile, is a generative approach really needed? Real traffic flow recordings can be extracted from the selected data source, and employed as if it was measured at the target location. On this premise, we conduct a study to analyze the benefits and downsides of using generative data methods against directly using the traffic recordings from the selected datasets as estimations for the target location. Finally, we examine the limitations of the presented road feature embedding selection method. The main contributions of our work can be summarized as follows:
\begin{itemize}[leftmargin=*]
    \item We design a set of graph-based features that characterize road segments, based on their topological and contextual characteristics. This road feature embedding allows spotting roads that are expected to have similar traffic patterns, without any need for traffic data collected in the location of interest. 
    \item The definition of such road feature embeddings heavily relies on intuition and domain-specific knowledge about the behavior of traffic flows under different circumstances. This clear match between knowledge and features contributes to the trustworthiness and actionability of traffic estimates predicted therefrom, which are essential in real-world traffic management decision making processes.
    \item We experimentally compare the performance of a similarity-based traffic estimation method relying on the aforementioned embeddings to that of a na\"ive  selection based on geographical proximity, where the nearest road segment with collected traffic data is used for estimation. Remarkably, the algorithmic transparency of the traffic estimation process using similarities computed over the aforementioned road embeddings adds to the overall reliability and trustworthiness of the methodology and produced estimates by traffic managers.
    \item We explore and comparatively evaluate different options to synthesize traffic data for the location of interest: 1) training a generative adversarial network (GAN) that models the multiple distributions of the selected traffic dataset; 2) clustering traffic samples according to certain criteria and provide as the output a representative traffic pattern; and 3) taking as estimation the real traffic data from the selected road segment. 
\end{itemize}

The rest of the paper is organized as follows: Section \ref{sec:previous works} reviews those works that analyze the topological and contextual influence on the traffic profile and delve into the capabilities of graph embedding methods for representing traffic networks. Section \ref{sec:materials} defines the features that conform the road feature embedding, how these embeddings are compared, the distinct generative approaches and the techniques employed for result assessment. The experiments to be conducted are described in Section \ref{sec:experimentation}. Here, the results steamed from such analyses are exposed and commented. To finish, Section \ref{sec:conclusions} summarizes obtained insights and outlines future research paths stemming from this work.

\section{Background}\label{sec:previous works}
In this section we provide a comprehensive view of the state-of-the-art concerning the main topic of this manuscript: the sensorless estimation of traffic profiles. Further arguments and motivations towards conducting this work are first addressed. Urban traffic modeling is a more challenging task regarding freeways (Subsection \ref{subsec:urban}). However, there is a lack of high-quality traffic flow data sources for these areas. Traffic simulators can provide synthetic urban traffic data. Even so, this data source type is associated to a sometimes overlooked concern: simulators do not consider the impact of certain aspects that influence urban traffic, such as the culture of distinct communities or the complex interplay between social events and traffic flow.

Finally, this section addresses the influence of network topology from two perspectives. The first analyzes the correlation between network topology and the empirically obtained traffic profiles (Subsection \ref{subsec:net design}). Being traffic flow a complex process, where multiple factors influence its behavior (e.g., culture, calendar, and weather), the network topology and the urban context appears to be one of them; one whose role is not completely clear yet. The second perspective is to represent the traffic network as a graph, a representational structure specifically designed to portray the similarities between road segments (Subsection \ref{subsec:graph}). Predictive models enhance their performance from the information within these networks' representations, to the point where most of published network-wide forecasting models in recent years are built upon this technology. 

\begin{figure*}[h]
    \centering
    \includegraphics[width=\textwidth]{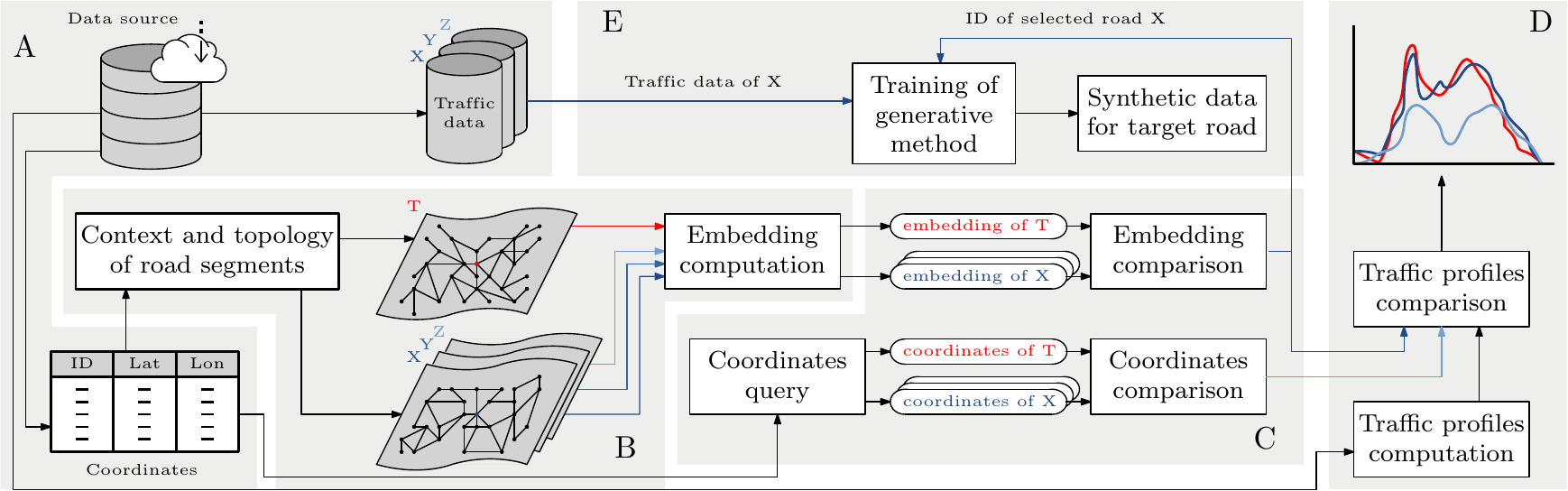}
    \caption{Workflow for this investigation. Capital black letters within each shaded region denote the index of the subsections in Section \ref{sec:materials} where the corresponding step of the workflow is discussed (e.g., A is discussed in Subsection \ref{sec:materials}.A). The red T stands for \emph{target road segment}, whereas X, Y and Z represent those locations considered as \emph{sensed road segments}. The colored dot represents the \emph{central node} of each ego graph.}
    \label{fig:workflow}
\end{figure*}

\subsection{Urban traffic and synthetic data} \label{subsec:urban}
Traffic behavior is more complex to model in cities, due to the multiple factors that can alter urban traffic \cite{manibardo2021deep}. This fact, combined with the low number of public urban RCD datasets, clarifies why not so long ago, urban road traffic data was scarcely studied \cite{vlahogianni2014short}.
It is not until recent years when research papers on this context have seen its number increased \cite{manibardo2021deep}. 
However, the majority of urban traffic-focused manuscripts were conducted over taxi or bike FCD. The traffic flow of the entire network can not be predicted from these kinds of datasets. Since FCD is gathered from individuals, only a fraction of the total flow can be modeled. Speed and travel time forecasting models can be learned from FCD, but when modeling the whole traffic flow of a road segment, RCD is preferred.


The disposal of synthetic data can provide benefits to the field, by giving access to urban traffic data of similar characteristics regarding a target location. Synthetic data is rooted in the idea of producing artificial samples following a statistical distribution \cite{brinkhoff2003generating}, which should be close to the real-world task to be modeled. Microscopic simulation tools, where the dynamics of each vehicle is modeled, such as SUMO \cite{krajzewicz2012recent}, VISSIM \cite{fellendorf2010microscopic}, CORSIM \cite{owen2000traffic}, and MATSIM \cite{w2016multi}, provide complex scenarios that allow to study not only traffic congestion, but also protocols for traffic light switching, emissions, energy consumption and so on \cite{lopez2018microscopic}. With such simulation tools, complex but still detailed flow traffic profiles for multiple road segments of a traffic network can be produced. Nevertheless, research can sometimes be heavily concentrated on the concept of traffic congestion, and therefore, it is mostly considered as influencing factors the weather, accidents reports or infrastructure change works. Simulation tools do not consider other fundamental aspects that concern the traffic profile. While freeway traffic can behave similarly under a fixed set of conditions, city traffic is highly influenced by additional factors, inherent to the urban areas. Events, road construction/restoration, etc., are difficult to include into simulations. The socio-cultural particularities of every major community at a city also models the traffic behavior. Still, it is hard to embed this knowledge within the configuration of simulation tools.


A methodology for generating synthetic traffic samples for non-sensed locations is presented in \cite{lana2021soft} to cover the above issues. The authors exploit the knowledge that can be learned from neighboring sensed road segments. Deep learning regression and GAN models are explored as data generation methods. It is a challenging task since there is not an explicit statistical distribution to be learned, due to the lack of real data at target locations. Likewise, traffic data exhibit a multi-modal nature, where not a single but multiple distributions or modes must be learned towards an accurate representation of data. In the context of traffic, these modes can be interpreted as the distribution followed by observed data under certain conditions (e.g., holidays and day of the week). Therefore, not only several distributions must be learned without the disposal of traffic data from the road segment of interest, but also the correct mode must be selected towards generating plausible synthetic data. This issue is addressed by using a set of conditions that groups traffic recordings according to the resemblance of their traffic patterns. Although the authors point out that mode selection can be improved by delving into the conditioning of generative systems, their main concern is to find a road segment with a similar traffic behavior. The generation systems will output divergent traffic patterns if the behavior of the selected roads is not close to the target location. Being exploratory research, a naïve criterion is adopted, where the closest ATRs available from the surroundings served as data sources. This approach entails speculation, where sometimes the selected data sources have similar traffic profiles, but in other cases traffic highly differs.


\subsection{Network design and traffic profiles}\label{subsec:net design}
The major challenge for a better design and usability of a traffic network is to understand the influence exerted by its structure. The topological design of a city map influences traffic behavior. Road occupation makes drivers choose distinct paths, pursuing a balance between selecting the shortest route and avoiding traffic congestion. In turn, simple street layouts facilitate for drivers to foresee other drivers' actions, so they can adapt their driving style without sudden braking or other similar abrupt movements. Nonetheless, with the inexorable expansion of the city population, the same applies to their infrastructure needs \cite{reilly2009bangalore}, where additional lanes and routes are appended to the existing layout, promoting traffic flow and hence, traffic congestion.

Some published works address this issue, seeking the relationship between traffic profiles and the design of the road itself. Wen et al. \cite{wen2017understanding} elaborate on the idea that numerous turning directions at crossroads can result in a hindrance for fluent traffic flow. By using collected data from the city of Taipei, Taiwan, they manage to identify the most congested segments of the metropolis (i.e., business districts and industrial areas).
On the same line, Wang et al. \cite{wang2018analyzing} conducted a novel study in the city of Shenyang, China, about the relationship between traffic and the proximity of certain Points of Interest (POI) (i.e., bus stations, schools, and hospitals). 
They could only demonstrate a correlation of traffic flow with the number of lanes. Despite these results, authors warned about the confidence of the presented conclusions, calling for more case studies over other cities towards a better understanding of the considered relationships. A whole research line stems from Geroliminis and Daganzo \cite{geroliminis2008existence}, where the first Macroscopic Fundamental Diagram (MFD) was obtained from a real environment (i.e., the city of Yokohama, Japan). The introduced MFD provides a description of the ideal network dynamic performance, drawing a curve that defines the critical point where an increase in vehicle density leads to a decrease in vehicle flow. As classified by Ambühl et al. \cite{ambuhl2021disentangling}, in later years distinct versions of the MFD have come out, but for the substance of this manuscript, the most relevant is what they define as theoretical upper bound MFD (uMFD). Just from the network topology and the traffic control policies, it is possible to simulate the uMFD, towards a better understanding of the theoretical traffic flow ceiling or, in other words, to foresee which is the maximum capacity of the network during an ideal driver behavior. In this line, several road network designs can be studied for selecting the one with higher capacity. Laval and Castrill\'on \cite{laval2015stochastic} proposed and stochastic approximation for computing the uMFD without the need for traffic flow measurements, just only from the block length and the traffic lights timings. A comparison with the empirically obtained uMFD from the city of Yokohama validates the method. Still, this technique does not have into consideration the particularities of other specific transport means (e.g., bus fleets). The latest advances in this investigation line are covered at the introduction of \cite{wong2021network} and \cite{sirmatel2021stabilization}, in the case of further reading needs.

\subsection{Traffic forecasting and graph representation}\label{subsec:graph}
As a subset of Machine Learning, deep learning \cite{lecun2015deep} has gained momentum in recent years, thanks to the impressive results obtained in several fields such as natural language processing \cite{landolt2021taxonomy} or computer vision \cite{hassaballah2020deep}. The excellent modeling capabilities of deep learning architectures expanded the competence of multiple research fields, though these improvements were temporally constrained to Euclidean data (i.e., data that can be expressed in a $m$-dimensional Euclidean space) \cite{darmochwal1991euclidean}.

Traffic forecasting is a problem that can be formulated in an Euclidean space. Traffic flow, speed or travel time are usually expressed as time series. A forecasting model receives as input the traffic state at times $\{t-n,\dots,t\}$ and outputs the expected traffic state at $t+h$, where $t$ is the current time instant, $n$ represents a time window of past traffic values, and $h$ is the forecasting horizon. 
The superior modeling capabilities of deep learning made authors concentrate their research efforts on solving traffic forecasting problems with this family of learning models. Different deep learning architectures have been proposed over the last years: from basic convolutional and recurrent networks to more complex and deep architectures such as those based on the attention mechanism \cite{vaswani2017attention} (a more comprehensive analysis of related efforts to date is provided in \cite{manibardo2021deep}). State-of-the-art architectures from other fields were applied to traffic forecasting, expecting similar levels of predictive performance. However, as demonstrated in \cite{manibardo2021deep}, deep learning architectures do not necessarily outperform conventional machine learning methods (e.g. ensemble learning). The traffic state at previous instants $\{t-n,\dots,t\}$ contains enough valuable information given the high persistence of the traffic time series to be predicted. Even a na\"ive model where the latest recorded traffic state value serves as the predicted traffic value can be shown to yield a sufficiently good forecast for low values of the horizon $h$. In summary, the feature learning capability of deep learning models is not differential for this specific application scenario, and poses further problems such as the interpretability of the knowledge captured by such black-box models once they have learned to forecast traffic \cite{arrieta2020explainable}.  

It was not until the upsurge of \emph{geometric deep learning} \cite{bronstein2017geometric} when multiple techniques arose towards developing new applications concerning non-Euclidean data: those tasks where data can be described as a graph \cite{harary2018graph} or as a manifold \cite{munkres2018analysis}.
As for the subject of this manuscript, traffic networks can be easily defined as graphs, where the road segments are the edges, and the crossroads are the nodes (although some authors prefer the opposite representation). This way, an abstraction of the traffic network can be defined as a graph $\mathcal{G} = (\mathcal{V}, \mathcal{E}, \textbf{A})$, where $\mathcal{V}$ is a set of $N$ nodes representing the junction of road segments, $\mathcal{E}$ is a set of edges symbolizing those road segments, and $\textbf{A}$ is a binary adjacency matrix in which element $a\textsubscript{i,j}$ represents the \emph{reachability} between nodes $i$ and $j$ of the traffic network: node $j$ can be reached from node $i$ if $a\textsubscript{i,j}=1$.

Bearing the above definitions in mind, graph neural networks (GNNs) are a type of deep learning architecture that is specifically designed to exploit the information represented by graph data structures \cite{wu2020comprehensive}. From the development of novel GNNs, researchers achieved to apply this technology towards traffic forecasting \cite{jiang2021graph}, as the traffic state (e.g., flow or mean speed) can be easily encapsulated as node features. This way, each node of the graph has a collection of features, composed by its traffic state of previous time instants. According to the reachability defined by $\textbf{A}$, the GNN can perform operations with not only the traffic features of one node but also merging the information of its neighbor nodes.

Given the ability of graph representations to condense the characteristics of a road network into a non-Euclidean space, the surroundings of a road segment should be able to be expressed as a collection of graph-based features that conforms to intuition and domain knowledge about behavioral patterns of traffic flows in urban networks with different topologies. For this reason, the road segment similarity finding method proposed in this work is based on features distilled from a graph representation which is intended to portray the topology and context of the neighboring.

\section{Materials and methods}\label{sec:materials}

It can be inferred from the literature review that a comparison between two roads of a traffic network can be performed by contrasting their topological features. Likewise, graphs have been proven to be a competent traffic network representation in previous literature. In this section, we exploit this information to create a method for comparing the similarity of two road segments without the need of traffic measurements.

Figure \ref{fig:workflow} represents the workflow of this investigation, which is designed to provide an overview of all the processes to be explained. First, available real-world traffic data is obtained from the data source at hand, along with the coordinates of every sensed and non-sensed road segment 
(Subsection \ref{subsec:representation}). These coordinates are used to compute a set of features which are intended to portray the context and topology of such locations: a road feature embedding (Subsection \ref{subsec:embedding}). By comparing these embeddings, a sensed road is selected (Subsection \ref{subsec:selection}). Similarly, the nearest sensed road is selected by inspecting the coordinates of every location (i.e., geographical selection method). The performance of both selection methods is analyzed by comparing the traffic profile of the target and selected roads (Subsection \ref{subsec:comparison}). Finally, three generative methods are introduced, towards analyzing the quality of the produced synthetic traffic data (Subsection \ref{subsec:generative}). Algorithm \ref{alg:methodology} summarizes the data and steps involved in a generic implementation of the proposed methodology for the sensorless estimation of road traffic profiles.

\begin{algorithm}
    \DontPrintSemicolon
	\SetKwIF{If}{ElseIf}{Else}{If}{}{ElseIf}{Else}{end}
	\SetKwFor{For}{For}{}{end}
	\KwIn{Coordinates $C$ of target $t$ and sensed road segments $\mathbf{s} =\{s_1,\ldots, s_n\}$, historic records of traffic flows $\mathbf{F}[s]$, ego-graph $e\mathcal{G}$, road feature embedding $\mathbf{E} = [f_1,\ldots, f_7]$ composed by features $f$, generative model $m$, and date of day to be predicted $d$
	}
	\KwOut{Traffic data estimation at target location}
	
	\For {$t$ and $\mathbf{s}$}{ 
	Compute $e\mathcal{G}$ according to $C$ \;
	Compute graph-based features $\mathbf{E_g}[e\mathcal{G}] = [f_1,\ldots, f_5]$, Expressions \eqref{eq:spbc} and \eqref{eq:tt2road}\;
	Obtain topological features $\mathbf{E_r}[C] = [f_6, f_7]$\;
	Arrange all features $\mathbf{E} = [\mathbf{E_g}[e\mathcal{G}], \mathbf{E_r}[C]]$\;
	Standardize $\mathbf{E}$\;
	}

	Find $s$ that minimizes $d(\mathbf{E}[t],\mathbf{E}[s])$, Expression \eqref{eq:euclidean distance}\;
	Train $m$ with $\mathbf{F}[s]$ \;
	Estimate traffic flow at the target location $t$ as $m[d]$\;
	\caption{Methodology for the sensorless estimation of road traffic profiles}
	\label{alg:methodology}
\end{algorithm}

\subsection{Traffic data and graph representation}\label{subsec:representation}
The traffic data for this investigation, is obtained from the city council of Madrid, Spain \cite{Madridddbb}. It provides access to multiple ATRs (i.e., inductive loops), that have collected traffic variables like flow, occupation or speed for several years. Specifically, the time span of the selected datasets for this investigation covers the 2018 and 2019 years. The coordinates of a total of 55 considered road segments is stored in a tabular file. Standard preprocessing techniques are applied to each traffic dataset, towards dealing with outliers or missing data, among other common data anomalies.

\begin{figure}[h!]
    \centering
    \includegraphics[width=\columnwidth]{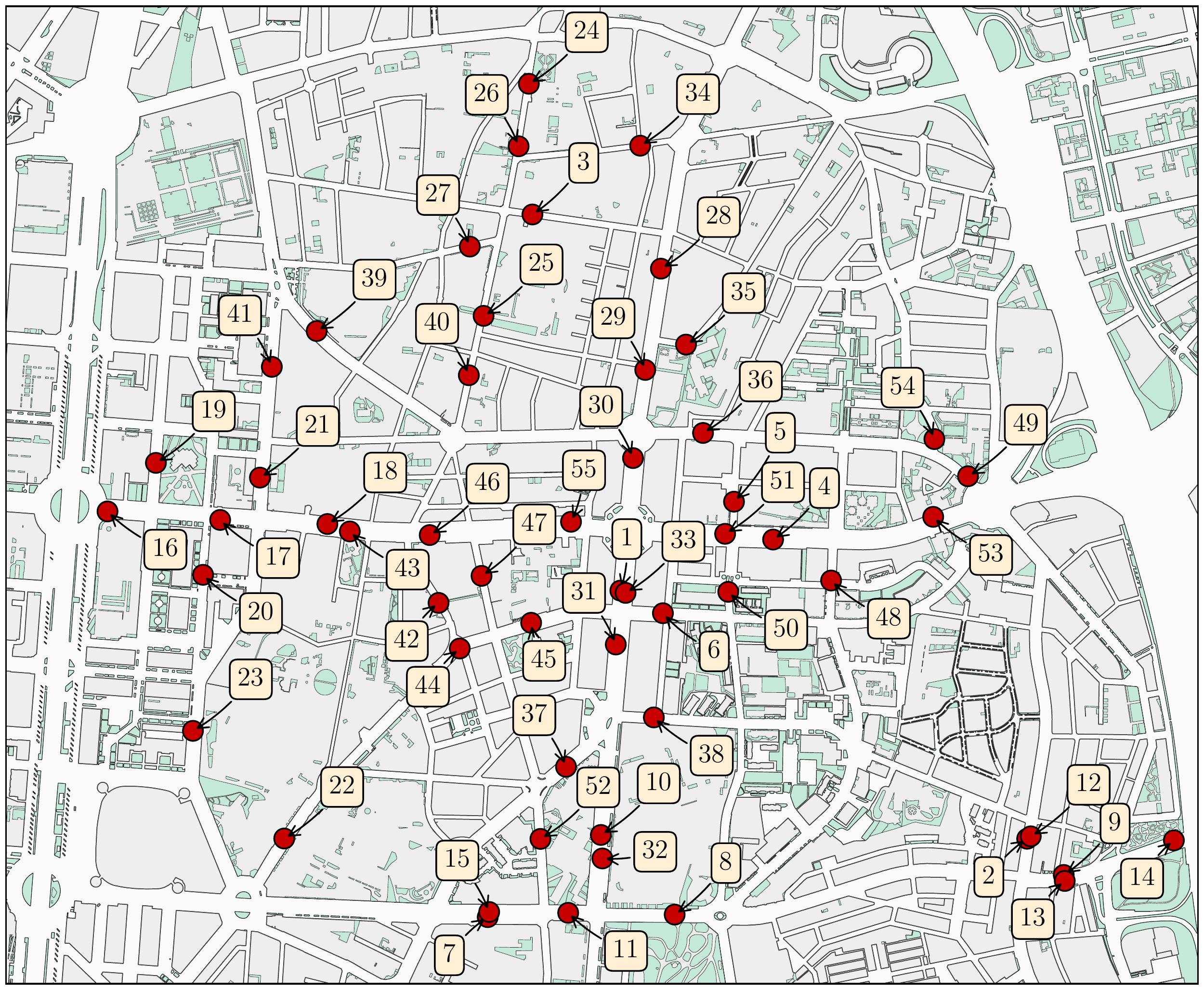}
    \caption{Location of the deployed sensors in the neighborhood of Chamart\'in in Madrid, Spain. Several road types are considered. Namely: secondary, tertiary, and residential roads, along with road links between distinct road types.}
    \label{fig:ATRs}
\end{figure}

The location of the ATRs is displayed in Figure \ref{fig:ATRs}. Here, it can be observed that all considered locations correspond to secondary, tertiary, or residential roads. Motorways and primary roads are excluded on purpose, as they are meant to be the infrastructure for in/out city journeys. The traffic behavior on these driveways is expected to be differentiated regarding the traffic flow of the lower-rank roads contemplated for this investigation.

The traffic network graph representation $\mathcal{G}$ is computed via the OSMnx Python package \cite{boeing2017osmnx}, which allows obtaining geospatial data from OpenStreetMap \cite{OpenStreetMap}. From the coordinates of each ATR, a directed graph is built within a 2km radius. Such distance has been established after a systematic search but must be long enough so every influential street is captured. This way, the road segments conform the graph edges $\mathcal{E}$, whereas the nodes $\mathcal{V}$ represent the crossroads. The road length, maximum speed, and travel time of the road (computed from the previous characteristics) are assigned as edge attributes for $\mathcal{E}$. Since the graph is obtained from the coordinates of an ATR, and this one is placed in a road segment, the focal point of the graph would be an edge. However, for the sake of an easier graph processing, the ATR should be placed as the central node of the network. To this end, the corresponding edge is split into two smaller segments which are, at the same time, connected to an artificial central node. In the following, the artificial node is considered as the focal point of the graph, which shares the coordinates of the ATR, and is ultimately referred to as \emph{central node}.

\subsection{Road feature embedding}\label{subsec:embedding}
A road feature embedding is designed to compare street segments. Some of the features that conform the embedding are based on a graph representation. However, the graph representation $\mathcal{G}$ is too wide to represent only the particularities of the target road. Therefore, an ego-graph $e\mathcal{G}$ \cite{sandfelder2021ego} is obtained by pruning all nodes in the original graph $\mathcal{G}$, that need more than $N$ hops to be reached from the central node. The parameter $N$ regulates the relevance of the surroundings during road segment comparisons. Low values for $N$ will make the system to focus on the topological aspects of the target street, while high $N$ values can make ego-graphs similar to each other, only because distant road segments regarding the central node are overlaid. This last case neglects the particular traits of the target road, which can produce non desirable resemblances. With these concerns in mind and after a grid search, a value of $N=5$ is set for this investigation. This parameter can be tuned according to the traffic map and feedback delivered by experts in traffic management with experience in the urban area under study. Complex traffic network designs might need higher values of $N$, since road segments are usually shorter, and hence small surfaces can be well represented by $e\mathcal{G}$. 
Conversely, the characterization of long road segments might require low $N$ values to avoid the overlap of similar graph representations.

From the ego-graph centered in the target location, a set of features is arranged. The NetworkX Python package \cite{hagberg2008exploring} provides the methods used to compute them. A first subset of such features is meant to express the \emph{centrality} of the central node and its neighbor nodes. As its name suggests, these metrics are intended to quantify the prominence of an agent (node or edge) in a network \cite{wasserman1994social}. This concept is motivated from the data networks domain, which often shares properties with traffic networks \cite{holme2003congestion}. Among the existing centrality measures available in the state-of-the-art \cite{valente2008correlated}, we have selected the \emph{shortest path betweenness centrality} ($SPBC$) which, for a node $v$, is the count of all pairs shortest paths that pass through it. Mathematically:
\begin{equation} \label{eq:spbc}
SPBC(v) = \sum_{s,t \in V} \frac{\sigma(s,t | v)}{\sigma(s,t)}
\end{equation}
where $\mathcal{V}$ is the set of considered nodes (neighbors of $v$), $\sigma(s,t)$ is the number of shortest $(s,t)$ paths, and $\sigma(s,t|v)$ is the total of those paths passing through node $v$, as defined in \cite{brandes2001faster}. 

Since in a graph representation crossroads serve as the edge split criterion, the streets represented in $\mathcal{E}$ are usually divided into multiple edges. Some long road segments are partitioned several times, while others with a reduced number of crossroads are represented by fewer, yet longer road segments. Upon this principle, the shortest path is computed by comparing the travel time assuming free-flow speed, instead of the number of hops (which is the default metric). From Expression \ref{eq:spbc}, the following metrics are computed:

\begin{enumerate}
    \item \textit{SPBC of the central node}: Higher values should correlate with high flow profiles.
    \item \textit{Maximum SPBC among neighbors}: It searches for other more essential nodes in $e\mathcal{G}$, regarding the central node.
    \item \textit{Median SPBC among neighbors}; It helps figure out the distribution of centrality, among nodes in $e\mathcal{G}$.
\end{enumerate}

Other centrality measures have been considered but excluded, as they do not reflect well the traffic profile of a road segment, or more precisely, it was intended to get rid of those features that could be ambiguous (road similarity or disparity could be argued with the same value). It is the case of \emph{degree}, which in the context of traffic networks is the number of links a crossroad has. Multiple connections to other road segments do not directly imply that a crossroad is heavily used, as many of these links can be scarcely frequented streets. In turn, a low-degree arterial road can be an essential milestone for numerous paths that go across the area. The \emph{closeness} of a node is the distance to all other connected nodes in the graph, or in other words, it measures how long it will take to spread information (or vehicles in this context), from the central node to all other nodes sequentially. While closeness is oriented to measure the broadcast capabilities of the nodes, it is a centrality measure focused on information networks. In contrast to data packages, vehicles are treated as individuals with no capacity for duplication. Instead of granting importance to the reachability of all other nodes, only certain points of the network should be interesting to be close to, as drivers are prone to searching for higher-rank road segments (e.g., primary roads). Departing from this idea, new embedding features are defined further on.

The neighborhood of Chamart\'in is enclosed by a primary road at the west, and by a motorway at the east (see Figure \ref{fig:ATRs}). In a large city like Madrid, the population usually travels in/out of the district through the fastest route, which generally compromises high-rank roads. The travel time towards reaching the closest primary road and/or motorway can elicit insightful features. This concept can be expressed as:
\begin{equation} \label{eq:tt2road}
TT(v,type) = \min \Delta(v,e_{type}) \forall e_{type} \in \mathcal{E},
\end{equation}
where $TT(v,type)$ is the expected travel time for a vehicle in node $v$, towards reaching the closest road of certain rank $type$ (in terms of travel time), and $\Delta(v,e_{type})$ is the expected travel time from $v$ to the closest node of a street segment $e$ of a certain rank $type$ belonging to a set of edges $\mathcal{E}$. 

This time, instead of using the ego-graph, the overall graph $\mathcal{G}$ is employed, as high rank roads could not fall into $e\mathcal{G}$. From Equation \ref{eq:tt2road} the following features are obtained:
\begin{enumerate}[resume]
    \item \textit{Travel time to the nearest motorway}: Motorways can be oriented to trips out of town.
    \item \textit{Travel time to the nearest primary road}: Primary roads can be oriented to trips to the city center.
\end{enumerate}

During rush hours, flow spikes can be narrow or wider depending on the overall flow. In the case of dormitory towns, most drivers go out of the city in the morning and return home in the evening, while major cities oppositely receive traffic. These previous features are tailored towards characterizing the events that impact traffic profile during rush hours.

Additionally to the features obtained from analyzing the graph, two extra features based on road characteristics are included for characterizing the road segments: the \emph{road type} and the \emph{number of lanes}. Each road type is encoded as an ascending scale where zero represents residential roads and the unit value corresponds to a motorway. Intermediate road types (namely tertiary, secondary, and primary) are encoded as in-between numbers, ensuring that all values are uniformly distributed. In contrast, the number of lanes is directly expressed by its original value. However, it is worth mentioning that although the NetworkX Python package provides methods for obtaining this road feature, there are some aggregation errors (i.e., merging the lanes of both directions), and missing values (generally in residential and tertiary single lane roads). Under this premise, each of the ATRs considered for this investigation has been manually inspected via Google Street View, to verify and rewrite the number of lanes if necessary. In those cases where the number of lanes varies, the predominant number of lanes across the entire length of the road segment is selected.


The above topological features are meant to give some notion about the traffic behavior in the studied road segments. Roads are classified in classes or ranks towards denoting their importance within the road network as a whole. Their structure, capacity, speed limits and even lane width varies between high and lower-rank roads. The number of lanes is just one specific parameter that defines the road structure. Nonetheless, it can produce more insights about the traffic flow regarding other road design features, such as the speed limit, which in the context of urban networks is most of the time equal or below 50 kph. Still, even if road networks are designed bearing in mind the expected traffic demand, real daily traffic profiles might not correlate with the expectations \cite{braess2005paradox}. Bearing this in mind, the last features that conform to the road feature embedding are defined as:

\begin{enumerate}[resume]
    \item \textit{Road type}: Higher rank road types are expected to portray higher daily traffic flow profiles. 
    \item \textit{Number of lanes}: Directly defines the road capacity and, henceforth the contemplated traffic flow.
\end{enumerate}

\begin{figure}[ht!]
    \centering
    \includegraphics[width=0.9\columnwidth]{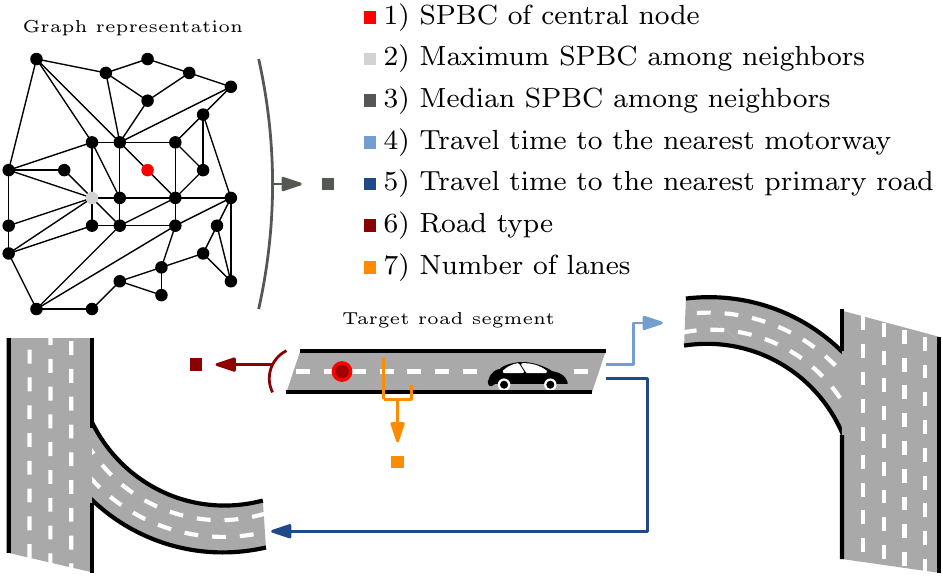}
    \caption{Graphical representation of the topological and contextual factors that conform the road feature embedding.}
    \label{fig:features}
\end{figure}

For the sake of a better understanding of the concepts beneath each topological and contextual feature, Figure \ref{fig:features} illustrates how such features are extracted from a road segment.

While some features (i.e. those based on centrality, road type, and the number of lanes) can be useful for any target area, features 4) and 5) are developed ad-hoc based on discussions held with experts in traffic analysis. The area considered in the case study (see Figure \ref{fig:ATRs}) is surrounded by a motorway and a primary road. These two arterials can be regarded as the major in- and out- traffic gates flowing into and out of the district, so their proximity to them (in terms of travel time) might be an interesting trait for characterizing road segments. Traffic managers may use their knowledge of local traffic dynamics even further for designing more features that represent other particular characteristics of the road segments, so the accuracy of the road feature embedding comparison can be enhanced. At this point we remark that this work is constrained by the available data of the target region. Nevertheless, future implementations can also benefit from accessing additional valuable information, such as the lane width or the existence of parking slots along the roadside.

\subsection{Selection of a sensed road segment}\label{subsec:selection}

Road feature embeddings are intended to portray the characteristics of the road segments. Therefore, two locations with a close traffic profile, should share similar values for the features that compose their respective embeddings. After a feature normalization, road feature embeddings can be represented as single points in a N-dimensional space. From this, by selecting the road feature embedding of the target road as origin, the Euclidean distance to any of the other road feature embeddings can be computed \cite{tabak2014geometry}:
\begin{equation} \label{eq:euclidean distance}
D(\mathbf{u},\mathbf{v}) = \sum_{n=1}^N \sqrt{|(u_n - v_n)|^2}  
\end{equation}
where $D(\mathbf{u},\mathbf{v})$ is the Euclidean distance between two N-dimensional vectors $\mathbf{u}$ and $\mathbf{v}$, whereas $u_n$ and $v_n$ are the features for the corresponding vector in position $n\in\{1,\ldots,N\}$. Since similar road feature embeddings should likewise have (according to our hypothesis) close values of their feature values, the minimal Euclidean distance among all available sensed road segments can be found. 

After selecting one road via the road feature embedding method, a second location is selected according to the coordinates of the ATRs and the target location. This way, the performance of both geographical and road feature embedding selection methods can be analyzed.

\subsection{Traffic profiles and selection performance}\label{subsec:comparison}
The main objective of the current manuscript is to present a method that allows to find similar roads in terms of traffic profile, without the need to compare the real traffic profiles between roads. However, for the sake of evaluating the performance of the road selection, the traffic profiles of both sensed and target roads must be set side by side and compared. Every road segment receives interactions from the surroundings, which uniquely condition its traffic profile, making perfect matches between target and selected locations unattainable. Still, as this method is intended to provide an alternative real traffic data source drawn from an analogous road for the target location, it is mandatory to pursue suchlike traffic profiles.

A traffic profile is a flow pattern that expresses the traffic behavior at a certain location of the traffic network. At a road segment, traffic flow varies with regard to the time of the day, the day of the week, holidays, and so on \cite{lana2016understanding}, making it arduous to condense all the traffic information in a single traffic pattern. Even so, a traffic profile must be computed in order to characterize distinct points of the city. Daily traffic flow metrics are more likely to be similar for two road segments that possess analogous traffic patterns and vice-versa. Henceforth, the comparison of traffic profiles for each of the 55 considered road segments is employed as a performance gauge for the proposed system. 

\begin{figure}[h]
    \centering
    \includegraphics[width=\columnwidth]{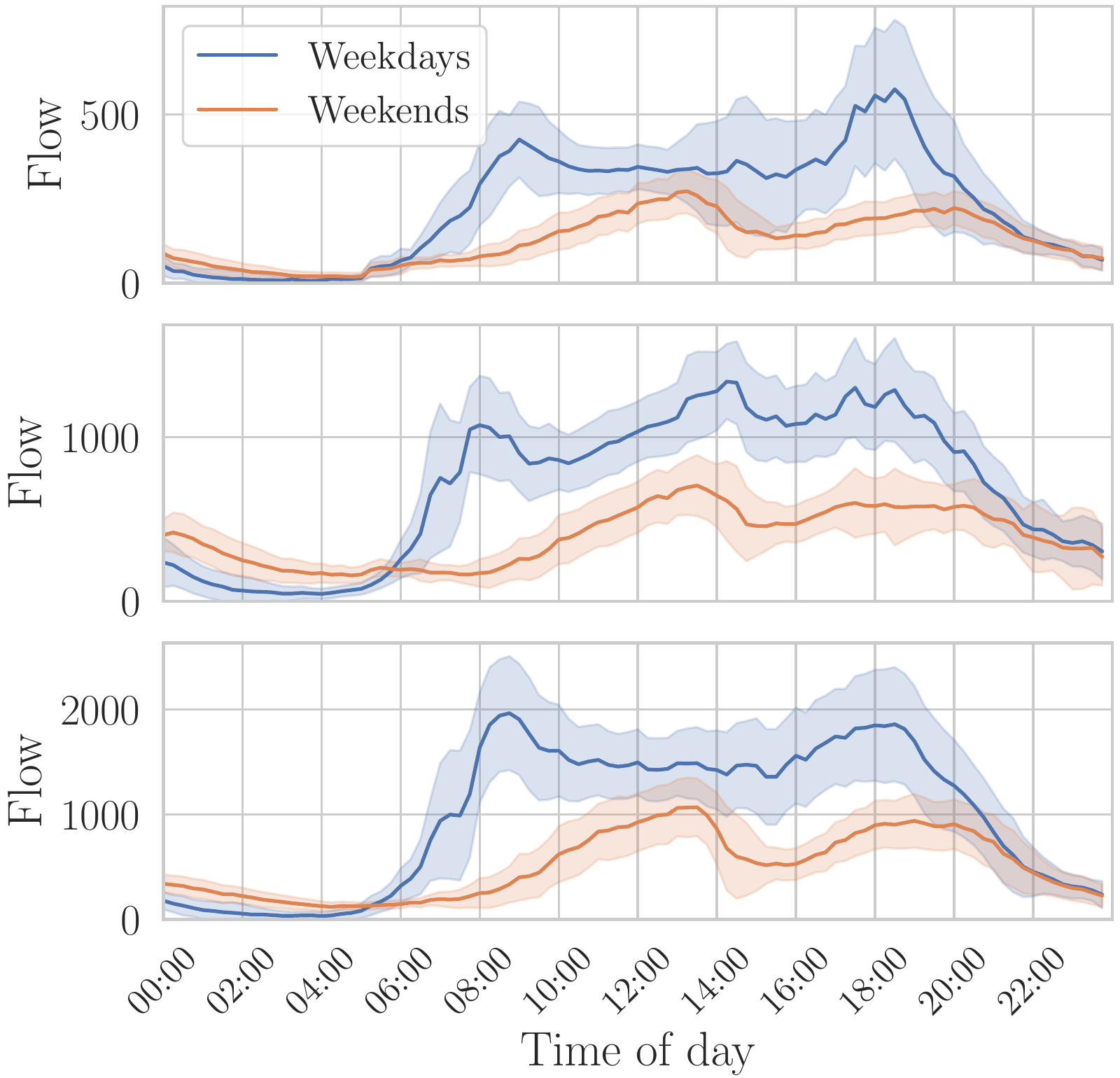}
    \caption{Traffic profiles computed from weekdays or weekends only, for three of the considered road segments. The continuous line represents the median flow value for each timestamp during a two-year period. Standard deviation is also displayed for each timestamp.}
    \label{fig:profiles}
\end{figure}

We propose to compute the median flow to obtain such traffic profiles. Since traffic highly differs between daily hours, the traffic flow for a certain timestamp is computed as the median of all recordings at that specific time. The median filters extreme flow values that can occur during special events such as Christmas, the beginning and return of the summer holidays, etc. The mean, on the other hand, would produce unrealistic traffic profiles, triggered by the aforementioned outliers. When it comes to filtering atypical values, only weekdays are considered for computing the traffic pattern, leaving out the traffic recordings from weekends. Traffic profile at weekends has fewer particularities than during weekdays, where usually the traffic profile is more restrained, as can be seen in Figure \ref{fig:profiles}. In addition to this, the weekends-only traffic patterns usually share a similar shape, to the point that with a proper scaling factor, the majority of these flow patterns could be obtained. By comparison, weekdays-only traffic patterns are more insightful towards characterizing distinct points of a traffic network, as they draw silhouettes that differ on the width and location of spikes. Additionally, Figure \ref{fig:profiles} also showcases that the standard deviation for weekends, obtained from all the flow recordings for a certain timestamp, is more uniform across the daytime. Putting all together, by excluding weekends in the calculation, more distinctive traffic patterns would be obtained. Therefore, the traffic profile is obtained exclusively from weekdays as:
\begin{equation} \label{eq:median}
y(r,t) = median(x(r,t,d))\:\:\text{$\forall d\in\mathcal{D}:\; d$ is weekday}
\end{equation}
where $y(r,t)$ is the traffic profile at a timestamp $t$ for a certain ATR $r$ whose traffic metrics are contained in dataset $\mathcal{D}$, $median(\cdot)$ denotes median statistic, and $x(r,t,d)$ denote the traffic flow available in day $d\in\mathcal{D}$ at a timestamp $t$.

\subsection{Methods for generating synthetic samples}\label{subsec:generative}
From the traffic data of the selected sensed road segment, synthetic samples can be generated for the target location. In the experimental framework of \cite{lana2021soft}, the authors select the two closest sensed road segments for generating data at the target location. This criterion does not ensure similar traffic profiles at the selected points of the traffic network, so a GAN is employed towards mixing the information of such locations. This way, a more stable solution is sought. On the contrary, the road feature embedding approach is designed to identify road segments with expected comparable traffic patterns. Under this premise, the first candidate to be chosen should be a more accurate substitute for the target location, regarding second and third proposals. Therefore, only one sensed road is employed for generating data. 

Within the context of this work, GAN based solutions could not be the best approach. The GAN based approach presented in \cite{lana2021soft} loses its purpose when a single traffic dataset conforms the data source, as there is not information to be combined. Several data generation approaches are suggested in this section, which are further analyzed considering their complexity (i.e., computational effort), and prediction performance. 

Traffic data usually follows several distributions or modes. One distribution could define how traffic behaves on workdays, whereas other distribution can model traffic under extreme weather conditions (e.g., heavy rain or snowfall). Thus, characterizing properly all modes of a sensed road segment is a complex task, which deserves its own research line \cite{lana2019adaptive}. Nevertheless, when generating traffic data, it is critical to select the correct distribution from which to compute the synthetic sample. Without identifying data distributions, a generation query could be answered with any plausible traffic profile from the desired location, which will provoke inaccurate predictions.

Operation modes have been reduced to 14, as per defined in \cite{lana2021soft}, obtained from the combination of weekdays and holidays. Since this criterion is merely based on calendar features, data can be labelled without the need of flow-based features. Every generation approach is adjusted using as training data the selected dataset from the road feature embedding system. The performance of each generative approach is then analyzed by employing the traffic recordings of the target location as test data. A traffic pattern is generated for all days in the test dataset (giving a total of 730 days). Henceforth, for each of the 55 considered target locations, 730 error metrics are computed, by comparing the real traffic with the generated traffic pattern.

The first generation approach revolves around training a conditional GAN \cite{mirza2014conditional}, conditioned with the 14 classes defined above. The goal is not to produce a top performance generation method, but to analyze distinct generation approaches. Under this premise, the \emph{approach} \protect\textcircled{\raisebox{-0.5pt}A} proposed in \cite{lana2021soft} has already showcased a prominent performance for this task, so it is selected as the generation method. A ReLU layer is added at the end of the generator model, so no negative traffic flow can be generated. The generative model receives as input the class of the desired day to be generated and a noise vector. Conceptually, this solution is based under the concept of learning several data distributions and taking as output a random sample of the selected mode. 

The second generation approach inherits concepts from the long-term estimation framework presented in \cite{manibardo2021change}. Here, traffic data is grouped by flow similarity into clusters. Each cluster disposes of a Representative Traffic Pattern (RTP), computed as the median of all the traffic samples within the cluster. From the value of certain calendar features, incoming days to be predicted are assigned to a cluster, and the aforementioned RTP serve as the traffic pattern that is going to be used as generated data. The framework is modified, so traffic data is grouped according to the aforementioned 14 classes. Given a date, that particular day is associated to one cluster. The RTP of the designated cluster is taken as the generated traffic pattern.

The last and third generation approach is the simplest and more easily interpretable by a user without any background on machine learning. If the target and selected road segments are considered similar, the traffic flow at a specific date should be similar in both locations. Under this premise, a Naïve Similarity-based Estimation (NSE) method is proposed, where the generated traffic samples for the target location are taken from the real traffic measurements at the selected road segment, for the same date. This approach is constrained to the disposal of traffic data for the date to be predicted. Hence, no traffic data can be generated for future dates, denying the prospective use of the system.

\section{Experiments and results}\label{sec:experimentation}

The design of our experimental setup departs from the findings and research directions reported in \cite{lana2021soft}, which emphasized that in order to improve the performance of generative methods for traffic data, a better criterion for selecting similar road segments should be conceived. The  selection of the right traffic dataset originates another dilemma, since synthetic data produced by a generative model could be no more accurate than taking as prediction the traffic record of the desired date from the selected dataset. Aiming to condensate these concerns, we formulate several research questions (RQ) to be answered with empirical evidence:
\begin{itemize}[leftmargin=*]
    \item RQ1: Can two road segments with similar traffic profiles be identified from a set of topological features?
    \item RQ2: Is there any relationship between road feature embedding similarity and the performance of our selection method?
    \item RQ3: Which is the best approach for generating synthetic data in the target location?
\end{itemize}

\begin{figure*}[ht!]
    \centering
    \includegraphics[width=\linewidth]{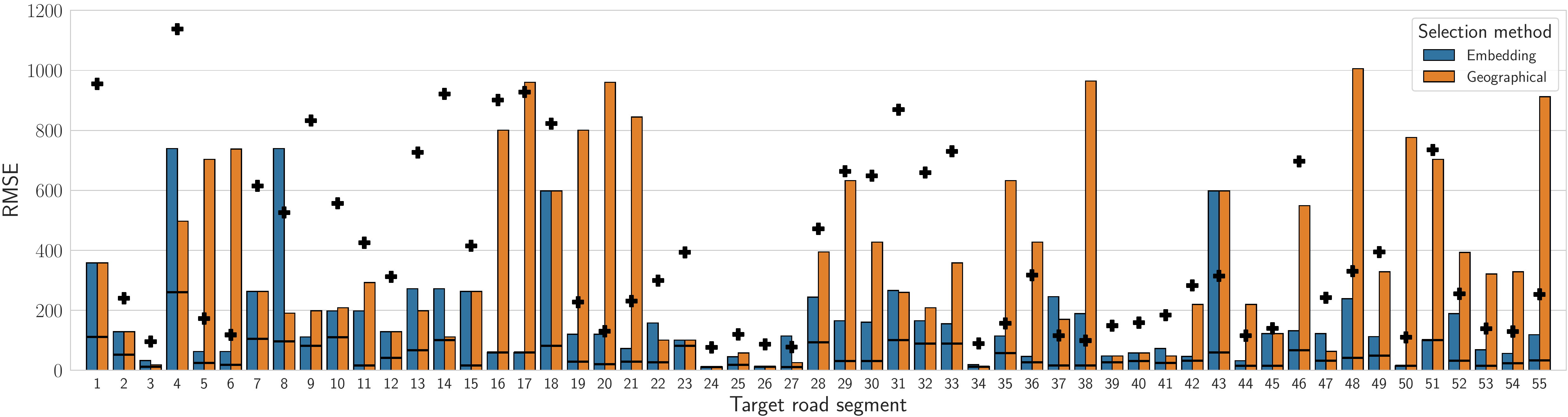}
    \caption{Performance of the selection methods for every considered road segment. Black lines showcase the lowest RMSE that would be obtained by selecting the most similar location among sensed ones if traffic measurements were available for comparison. Black crosses denote the mean flow value for the target road (considering only weekdays).}
    \label{fig:rmse}
\end{figure*}

Towards assessing quantitatively the performance of the different traffic generation methods, error metrics must be computed. This can be achieved by comparing the traffic profiles of two road segments. The Root Mean Square Error (RMSE) has been chosen as error metric, due to its explainability. Obtained deviations can be interpreted as the number of vehicles that might be under/over predicted. Given a desired traffic pattern, the RMSE measures the performance of the selection system:
\begin{equation} \label{eq:rmse}
RMSE(\mathbf{y},\mathbf{\widehat{y}}) = \sqrt{\frac{1}{N} \sum_{i}^{N} (y_i-\widehat{y}_i)^2},
\end{equation}
where $RMSE(\mathbf{y},\mathbf{\widehat{y}})$ is the RMSE computed over a traffic profile $\mathbf{y}$ from a target location and a traffic profile $\mathbf{\widehat{y}}$ from a sensed point of the network, and $N$ is the total number of samples that conform the traffic profiles over which the error is computed.

Being RMSE a scale dependent error metric, its normalized version or nRMSE is also employed in this section. There is no single normalization criterion in the literature; as such, commonly adopted options imply dividing the measure by the mean, standard deviation or the range of the data. The chosen approach is to normalize the RMSE by the mean flow of all weekdays from target location. In the context of this work, the standard deviation expresses how much the flow can vary among distinct days and conditions, so a normalization using this metric will be more punishing for those road segments with a fluctuating flow. Similarly, the range of data is also discarded towards normalizing the obtained error. The majority of roads have certain time periods where traffic flow is close to zero (i.e., early morning hours). This entails that the range of data is almost equal to the maximum flow value of the traffic profile. By normalizing the RMSE with the mean flow value, extreme values are smoothed, producing a more robust metric.

Before proceeding with the discussion of the obtained results, we note that, for the sake of reproducibility, all datasets in use, source code and results have been made available at \url{https://github.com/Eric-L-Manibardo/Road_embedding}.

\subsection*{RQ1: Can two road segments with similar traffic profiles be identified from a set of topological features?}
As previously introduced, the performance metrics of the road segment selection methods are obtained by comparing traffic profiles. For each target road, two performance metrics are calculated: one for the embedding selection method; a second one for the geographical selection method. Both metrics can be directly compared since they are computed using the same traffic profile as the ideal generated pattern. Additionally, the RMSE that would be obtained from the best fitting traffic profile among sensed road segments is also considered. This metric is intended to gauge the lower boundary error that could be obtained. To conclude, the mean flow value of all weekdays (since traffic profile calculation also excludes weekends) of the target road, portrays an upper boundary error. Putting all this together, Figure \ref{fig:rmse} presents the obtained results.

From a total of 55 considered road segments, comparing their road feature embeddings produces a better selection for 30 cases. The geographical criterion excels our proposed system in 12 cases. For the rest of the analyzed target roads, both selection methods output the same location (13 draws). Furthermore, a closer look reveals that the selection system based on road feature embedding produces more restrained errors, whereas the average committed error is far greater for the geographical method. The nearest sensed road segment does not have to share any topological relationship regarding the target location (e.g., road type or number of lanes). 

Figure \ref{fig:rmse} also includes the mean flow value for each target road segment, computed from weekdays only (as traffic profiles are computed from the same date range). Although there is not a consensus that relates the average traffic flow with the minimum accuracy for a system to be appealing, these values help to showcase the dimensions of each scenario. The same committed error has a different impact on the quality of the selection. When comparing traffic profiles, a difference of 50 vehicles is meaningful for a 100 average flow road. However, if the mean traffic flow is around 1000 vehicles, the performance of the system would be acceptable. 

If traffic measurements were available at target location, the black horizontal lines denote the error that would be obtained by selecting the road segment with the most similar traffic profile. Therefore, it is the lowest error that can be obtained with the current set of sensed road segments. This metric helps to contextualize the performance of the selection criteria. If the best selection performance that can be obtained from the current set of sensed locations is not acceptable, it means that either the target location has a peculiar traffic profile or more sensors should be deployed towards collecting traffic data at an analogous location. While the disposal of more sensed locations can enhance the performance of the selection method, the key point should be to gain the ability to spot those road segments of the network with unusual traffic profiles. This way, only a representative road segment for each type of traffic profile would be needed to be sensed, opening the gates to more restricted budget plans.

\subsection*{RQ2: Is there any relationship between road feature embedding similarity and the performance of our selection method?}

Being the Euclidean distance the criterion for searching the most similar road feature embedding, it is worth studying if there is any kind of relationship between this metric and the performance of the system. In order to extract insights from comparing the results of every road segment, a scale independent metric is needed. Therefore, the performance of each selection procedure is measured by the nRMSE score.

Along with the embedding selection method, we also analyze the embedding similarity that would be obtained from the locations with the most similar traffic profile (i.e., those that produce the lowest RMSE possible). By contrast, the geographical selection method is not considered for this topic, as this criterion is unpredictable: the closest sensed location does not have to possess a similar road feature embedding.

\begin{figure}[b]
    \centering
    \includegraphics[width=\columnwidth]{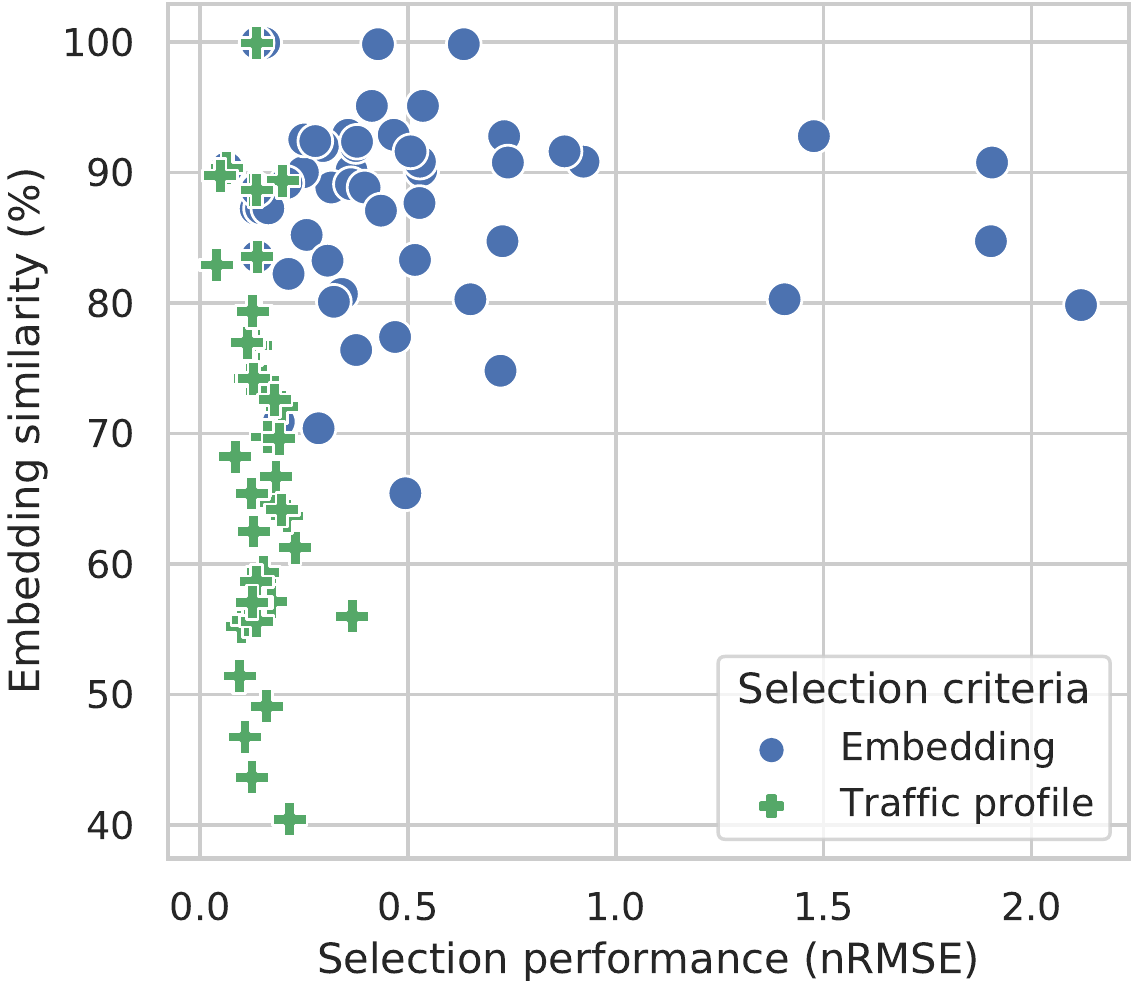}
    \caption{Road feature embedding similarity against selection performance for each of the analyzed road segments. In green, the selected location is the one with the most similar embedding. In blue, the selected location is the one with the most similar traffic profile (so traffic measurements at target location would be needed for this comparison).}
    \label{fig:euclidean}
\end{figure}

Road feature embeddings are compared by computing the Euclidean distance between a couple of embeddings. Since the seven considered features (see Figure \ref{fig:features}) are normalized to the range (0,1), from Expression \eqref{eq:euclidean distance} the maximum Euclidean distance can be computed as $\sqrt{7\cdot(0+1)^2}$. This way, two identical road feature embeddings would produce an Euclidean distance of zero, whereas the most dissimilar couple of road feature embeddings would be at an Euclidean distance of $\sqrt{7}$. From this, the embedding similarity can be expressed as a percentage.

Figure \ref{fig:euclidean} shows the relationship between selection performance and embedding similarity. In addition to the road segments with the most similar road feature embedding, for each target location, the road feature embeddings of the sensed locations with the most similar traffic profiles are also represented. 

No correlation between embedding similarity and selection performance is found. On one hand, the performance of the road feature embedding selection method is not optimal. There are a few cases where even with an embedding similarity above 80\% the selection performance is above the unit (i.e., RMSE above the mean flow of target location). On the other hand, the selection performance of the traffic profile selection criterion demonstrates that for the majority of considered target locations there is a sensed road segment with a comparable traffic pattern. However, it should be point out that some sensed locations exhibit a low embedding similarity while still producing a high selection performance. Therefore, analogous traffic profiles can exist at road segments of distinct contextual and topological characteristics. Under this premise, optimal selection performance could not be possible to achieve for every considered target location, as unrelated roads can also produce a similar flow.

To finish this second discussion, it is worth pointing out those cases with an almost identical road feature embedding (i.e., similarity near to 100\%). As some sensors are deployed geographically close to each other, their graph representation is similar. Road type and number of lanes is also shared. Sometimes, there are two sensed road segments with a direct connection between them. In other cases, both directions of the same avenue are present in the data collection. Sensors that share most of their context and surrounding network, also share most of their embedding feature values. This fact might produce sub-optimal road selections, where almost identical points of the network have distinct traffic behaviors.
\begin{figure}[h!]
    \centering
    \includegraphics[width=\columnwidth]{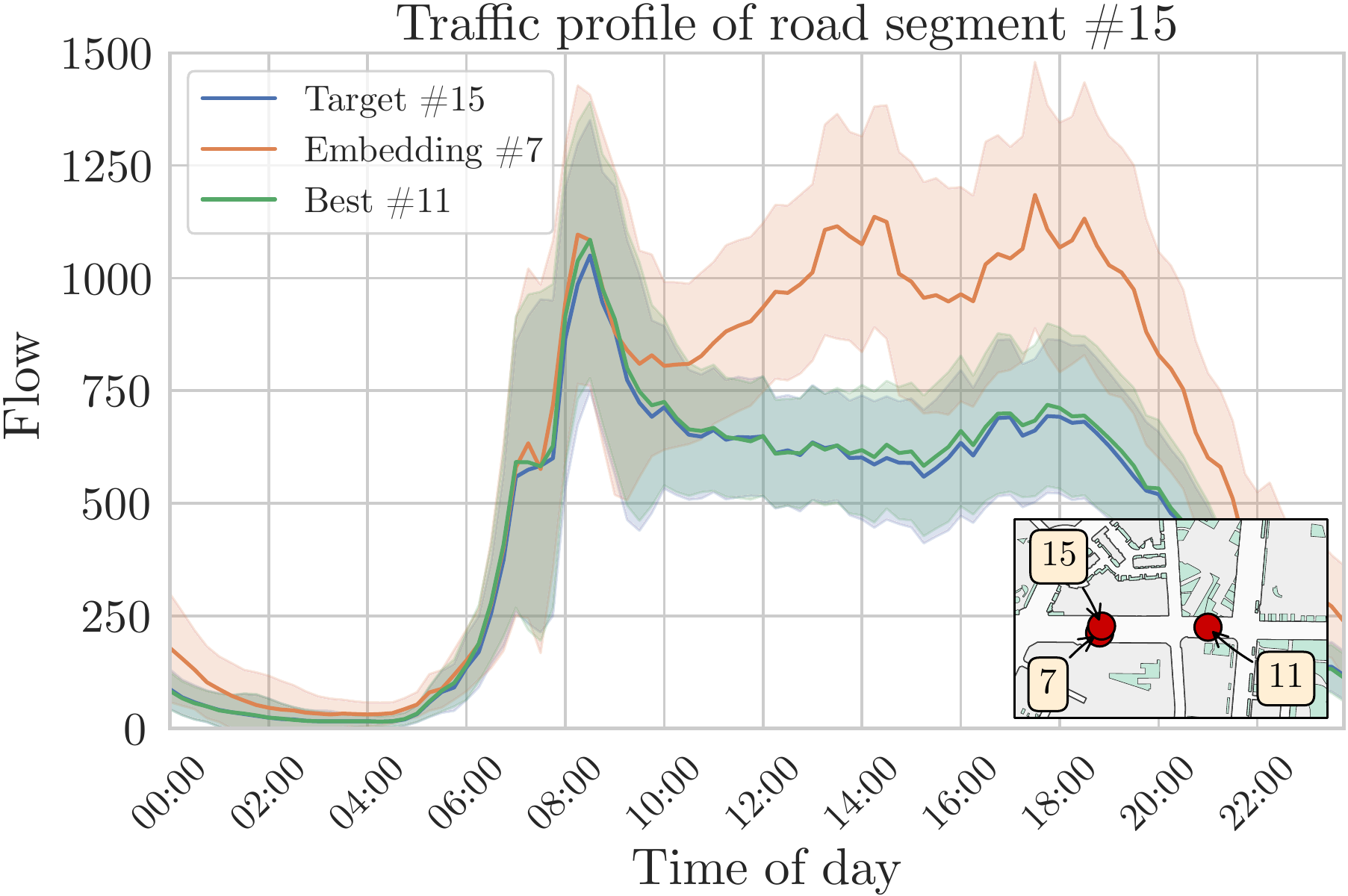}
    \caption{Comparative of traffic profiles. In blue, the target road. In orange, the traffic pattern of the most similar road according to its road feature embedding. In green, the best fitting traffic profile among sensed locations.}
    \label{fig:atr15}
\end{figure}

Figure \ref{fig:atr15} illustrates the above issue. Road segment $\#15$ and $\#7$ are placed in the same avenue, just in opposite directions (see Figure \ref{fig:ATRs}). As can be appreciated however, their traffic profiles highly differ after the morning rush hour. Being both primary roads, $\#15$ heads up to the city center, whereas $\#7$ conducts traffic towards the outskirts. Madrid is a city that receives daily traffic from the surrounding towns, in the form of workers going to complete a work shift. They leave the city center when returning home, which is why a high flow is maintained at $\#7$ until evening. Among sensed locations, the road segment $\#11$ is located in the same avenue and direction than $\#15$, producing an analogous traffic pattern (colored in green). Clearly, $\#11$ is a good representation of the traffic flow of $\#15$. Still, the road feature embedding of $\#7$ is closer to $\#15$, which is why a high error is obtained.

Nevertheless, in the context of this investigation the target location is not sensed, so no traffic measurements are available at that point of the traffic network. Without traffic data, only topological and contextual data can be used as criteria for selecting one sensed road. From the example of Figure \ref{fig:atr15} it could be justified the need of an additional feature towards representing if a sensed road shares avenue and direction with the target location. However, road segments are connected by intersections that can highly alter the traffic profile (both adding or subtracting flow). Without any information of the influence of such intersections the previous statement is not straightforward. This is the reason for not modeling this aspect in the road feature embedding.


\subsection*{RQ3: Which is the best approach for generating synthetic data in the target location?}

For each target location, traffic data is generated for a total of 730 days from the 2018-2019 year period. This process is replicated for the three considered generation approaches: \emph{GAN}, \emph{Cluster} and \emph{NSE}. Synthetic samples are compared to the real traffic flow measurements, thus obtaining a set of error scores. 

Table \ref{tab: generation} shows the mean and standard deviation of the 730 error values for each target location and generative approach. The normalized version of the RMSE allows comparing performance between considered road segments. Figure \ref{fig:rmse} represents the selection performance but can also be interpreted as an estimation of the generative performance for the same time period (if the selection error is normalized by the mean flow). The obtained normalized selection performance is close to the best generative performance obtained for the same location. This fact demonstrates the viability of weekday-only traffic profiles for characterizing road segments. However, since the computed traffic profiles are smoothed representations of the traffic behavior (i.e., no weekends or holidays are represented), the particularities of the daily traffic produce slightly worse performance results for the generative approaches.

High standard deviation values are reported concerning the mean performance error. This aspect is justified both by the size of the traffic datasets and by the analyzed task itself. Two whole years compose the test holdout. Special days like holidays and summer vacations can highly differ between the selected road segment and the target location, even if during the rest of the year they share a similar traffic profile. Likewise, not prediction methods are analyzed, but generative approaches instead. As synthetic data is not built from the traffic recordings of the target location, higher error metrics should be expected regarding short-term forecasting methods where the input features of the model is the past traffic state at target location.

Due to the high standard deviations exposed above, we inspect further the statistical significance of the differences found between the GAN, Cluster and NSE approaches by ranking them considering the outcomes of a hypothesis test. Specifically, for every target road segment, we perform a global Friedman test \cite{friedman1940comparison} for repeated measurements to ascertain whether any significant differences exist over the RMSE results obtained with every approach in comparison. If such significance holds at level $\alpha=0.05$, a Nemenyi post hoc test for unreplicated blocked data \cite{nemenyi1963distribution} is performed over the RMSE results of every pair of generative approaches, so that wins, losses and ties among the three comparison counterparts can be resolved taking into account the statistical significance of the differences in the means of the results.
\\

\begin{table}[ht]
\centering
    \begin{adjustbox}{max width=\columnwidth}
    \begin{threeparttable}
        \caption{Performance results for considered generative approaches \\ for every target road segment.}
        \label{tab: generation}
        \begin{tabular}{rccccc} 
         \toprule
         \multirow{1}{*}{}  &
         \multicolumn{1}{c}{\textbf{GAN}} &
         \multicolumn{1}{c}{\textbf{Cluster}}  &
         \multicolumn{1}{c}{\textbf{NSE}} &
         \multicolumn{1}{c}{\textbf{Road type}} &
         \multicolumn{1}{c}{\textbf{Best}}
         \\
         
         \cmidrule(lr){2-2}
         \cmidrule(lr){3-3}
         \cmidrule(lr){4-4}
         \cmidrule(lr){5-5}
         \cmidrule(lr){6-6}
         
         \textbf{1} & 0.50 $\pm$ 0.17	& 0.44 $\pm$ 0.18	& \textbf{0.38 $\pm$ 0.16} & Secondary & NSE \\
         \textbf{2} & 0.78 $\pm$ 0.31	& \textbf{0.59 $\pm$ 0.21}	& 0.61 $\pm$ 0.22 & Tertiary & Cluster\\
         \textbf{3} & 0.60 $\pm$ 0.18	& \textbf{0.51 $\pm$ 0.18}	& \textbf{0.49 $\pm$ 0.13} & Residential & Cluster, NSE\\
         \textbf{4} & \textbf{0.60 $\pm$ 0.19}	& 0.66 $\pm$ 0.21	& 0.64 $\pm$ 0.17 & Secondary & GAN\\
         \textbf{5} & 0.59 $\pm$ 0.21	& 0.55 $\pm$ 0.22	& \textbf{0.49 $\pm$ 0.17} & Residential & NSE\\
         \textbf{6} & 0.70 $\pm$ 0.28	& \textbf{0.66 $\pm$ 0.23}	& 0.71 $\pm$0.25 & Residential & Cluster\\
         \textbf{7} & 0.53 $\pm$ 0.18	& 0.49 $\pm$ 0.18	& \textbf{0.45 $\pm$ 0.14} & Secondary & NSE\\
         \textbf{8} & \textbf{1.31 $\pm$ 0.30}	& 1.42 $\pm$ 0.40	& 1.40 $\pm$ 0.38 & Secondary & GAN\\
         \textbf{9} & 0.43 $\pm$ 0.12	& 0.30 $\pm$ 0.16	& \textbf{0.21 $\pm$ 0.11} & Secondary & NSE\\
         \textbf{10} & 0.50 $\pm$ 0.19	& 0.45 $\pm$ 0.18	& \textbf{0.37 $\pm$ 0.15} & Secondary& NSE\\
         \textbf{11} & 0.70 $\pm$ 0.26	& 0.56 $\pm$ 0.25	& \textbf{0.48 $\pm$ 0.20} & Secondary & NSE\\
         \textbf{12} & 0.53 $\pm$ 0.18	& 0.51 $\pm$ 0.19	& \textbf{0.46 $\pm$ 0.17} & Tertiary & NSE\\
         \textbf{13} & 0.54 $\pm$ 0.16	& 0.44 $\pm$ 0.19	& \textbf{0.44 $\pm$ 0.14} & Secondary & NSE\\
         \textbf{14} & 0.44 $\pm$ 0.17	& 0.39 $\pm$ 0.14	& \textbf{0.34 $\pm$ 0.11} & Secondary & NSE\\
         \textbf{15} & 0.78 $\pm$ 0.26	& 0.69 $\pm$ 0.26	& \textbf{0.66 $\pm$ 0.21} & Secondary & NSE\\
         \textbf{16} & 0.40 $\pm$ 0.15	& 0.29 $\pm$ 0.16	& \textbf{0.14 $\pm$ 0.10} & Secondary & NSE\\
         \textbf{17} & 0.42 $\pm$ 0.12	& 0.28 $\pm$ 0.13	& \textbf{0.14 $\pm$ 0.10} & Secondary & NSE\\
         \textbf{18} & 0.75 $\pm$ 0.21	& 0.75 $\pm$ 0.22	& \textbf{0.74 $\pm$ 0.18} & Secondary  & NSE\\
         \textbf{19} & \textbf{0.52 $\pm$ 0.16}	& 0.57 $\pm$ 0.20	& \textbf{0.55 $\pm$ 0.15} & Residential  & GAN, NSE\\
         \textbf{20} & 0.97 $\pm$ 0.27	& \textbf{0.95 $\pm$ 0.29}	& 0.95 $\pm$ 0.27 & Residential & Cluster\\
         \textbf{21} & 0.56 $\pm$ 0.15	& 0.43 $\pm$ 0.18	& \textbf{0.41 $\pm$ 0.16} & Tertiary & NSE\\
         \textbf{22} & 0.65 $\pm$ 0.23	& 0.59 $\pm$ 0.25	& \textbf{0.57 $\pm$ 0.21} & Tertiary & NSE\\
         \textbf{23} & 0.52 $\pm$ 0.29	& 0.45 $\pm$ 0.30	& \textbf{0.40 $\pm$ 0.31} & Tertiary & NSE\\
         \textbf{24} & 0.73 $\pm$ 0.26	& 0.45 $\pm$ 0.21	& \textbf{0.26 $\pm$ 0.16} & Tertiary & NSE\\
         \textbf{25} & 0.60 $\pm$ 0.21	& 0.50 $\pm$ 0.20	& \textbf{0.40 $\pm$ 0.14} & Tertiary & NSE\\
         \textbf{26} & 0.56 $\pm$ 0.24	& 0.49 $\pm$ 0.24	& \textbf{0.23 $\pm$ 0.14} & Tertiary & NSE\\
         \textbf{27} & 1.97 $\pm$ 0.30	& \textbf{1.82 $\pm$ 0.29}	& 1.87 $\pm$ 0.39 & Residential & Cluster\\
         \textbf{28} & 0.65 $\pm$ 0.19	& \textbf{0.59 $\pm$ 0.21}	& \textbf{0.53 $\pm$ 0.13} & Secondary & Cluster, NSE\\
         \textbf{29} & 0.48 $\pm$ 0.17	& \textbf{0.35 $\pm$ 0.10}	& \textbf{0.33 $\pm$ 0.11} & Secondary & Cluster, NSE\\
         \textbf{30} & 0.48 $\pm$ 0.18	& \textbf{0.34 $\pm$ 0.13}	& \textbf{0.31 $\pm$ 0.13} & Secondary & Cluster, NSE\\
         \textbf{31} & 0.44 $\pm$ 0.14	& 0.40 $\pm$ 0.17	& \textbf{0.30 $\pm$ 0.15} & Secondary & NSE\\
         \textbf{32} & 0.44 $\pm$ 0.12	& 0.42 $\pm$ 0.15	& \textbf{0.34 $\pm$ 0.12} & Secondary & NSE\\
         \textbf{33} & 0.39 $\pm$ 0.10	& 0.37 $\pm$ 0.13	& \textbf{0.27 $\pm$ 0.09} & Secondary & NSE\\
         \textbf{34} & 0.50 $\pm$ 0.12	& 0.45 $\pm$ 0.16	& \textbf{0.36 $\pm$ 0.12} & Residential & NSE\\
         \textbf{35} & \textbf{0.85 $\pm$ 0.20}	& 0.85 $\pm$ 0.20	& \textbf{0.84 $\pm$ 0.18} & Residential & GAN, NSE\\
         \textbf{36} & 0.50 $\pm$ 0.18	& 0.40 $\pm$ 0.15	& \textbf{0.36 $\pm$ 0.18} & Tertiary & NSE\\
         \textbf{37} & 2.22 $\pm$ 0.61	& \textbf{2.12 $\pm$ 0.51}	& 2.25 $\pm$ 0.70 & Tertiary & Cluster\\
         \textbf{38} & 2.36 $\pm$ 0.51	& \textbf{1.94 $\pm$ 0.43}	& 2.00 $\pm$ 0.53 & Residential & Cluster\\
         \textbf{39} & 0.80 $\pm$ 0.17	& \textbf{0.46 $\pm$ 0.16}	& 0.49 $\pm$ 0.25 & Residential & Cluster\\
         \textbf{40} & 0.56 $\pm$ 0.22	& 0.52 $\pm$ 0.20	& \textbf{0.45 $\pm$ 0.16} & Tertiary & NSE\\
         \textbf{41} & 0.66 $\pm$ 0.19	& 0.52 $\pm$ 0.22	& \textbf{0.51 $\pm$ 0.20} & Tertiary & NSE\\
         \textbf{42} & 0.53 $\pm$ 0.22	& \textbf{0.43 $\pm$ 0.22}	& \textbf{0.40 $\pm$ 0.21} & Tertiary & Cluster, NSE\\
         \textbf{43} & \textbf{1.99 $\pm$ 0.65}	& 2.03 $\pm$ 0.39	& 2.06 $\pm$ 0.50 & Secondary & GAN\\
         \textbf{44} & 0.61 $\pm$ 0.56	& 0.58 $\pm$ 0.58	& \textbf{0.55 $\pm$ 0.58} & Residential & NSE\\
         \textbf{45} & 0.98 $\pm$ 0.27	& 0.90 $\pm$ 0.26	& \textbf{0.89 $\pm$ 0.24} & Residential & NSE\\
         \textbf{46} & 0.48 $\pm$ 0.15	& 0.32 $\pm$ 0.14	& \textbf{0.27 $\pm$ 0.10} & Secondary & NSE\\
         \textbf{47} & \textbf{0.52 $\pm$ 0.17}	& 0.56 $\pm$ 0.18	& \textbf{0.52 $\pm$ 0.14} & Residential & GAN, NSE\\
         \textbf{48} & \textbf{0.84 $\pm$ 0.25}	& 0.85 $\pm$ 0.24	& \textbf{0.84 $\pm$ 0.22} & Residential & GAN, NSE\\
         \textbf{49} & 0.45 $\pm$ 0.17	& 0.40 $\pm$ 0.17	& \textbf{0.29 $\pm$ 0.12} & Tertiary  & NSE\\
         \textbf{50} & 0.54 $\pm$ 0.24	& \textbf{0.44 $\pm$ 0.26}	& 0.62 $\pm$ 0.59 & Residential & Cluster\\
         \textbf{51} & 0.41 $\pm$ 0.13	& 0.28 $\pm$ 0.15	& \textbf{0.21 $\pm$ 0.13} & Secondary & NSE\\
         \textbf{52} & \textbf{0.74 $\pm$ 0.20}	& 0.77 $\pm$ 0.24	& \textbf{0.75 $\pm$ 0.20} & Residential & GAN, NSE\\
         \textbf{53} & 0.66 $\pm$ 0.20	& \textbf{0.58 $\pm$ 0.19}	& \textbf{0.58 $\pm$ 0.15} & Secondary & Cluster, NSE\\
         \textbf{54} & 0.61 $\pm$ 0.22	& \textbf{0.52 $\pm$ 0.22}	& \textbf{0.50 $\pm$ 0.19} & Residential & Cluster, NSE\\
         \textbf{55} & \textbf{0.56 $\pm$ 0.17}	& 0.58 $\pm$ 0.18	& 0.56 $\pm$ 0.15 & Residential & GAN\\
         \bottomrule
        \end{tabular}
        \begin{tablenotes}
          \item Note: displayed results are rounded due to space constraints. For performance assessment, all decimals are considered.
        \end{tablenotes}
  \end{threeparttable}
  \end{adjustbox}
\end{table}

The NSE approach dominates the benchmark, where for 78\% of the analyzed cases, is declared as the best by the Nemenyi test. Furthermore, the standard deviation denotes less statistical dispersion for the committed error. Even so, results are close among the three considered generation approaches. Model complexity and computational resources might be properties of concern for selecting one approach over the others. Coincidentally, performance decrease as the complexity of the generation approaches increases. In line with this, the inconveniences of the NSE method are twofold: 1) the need for traffic measurements for the desired date to be generated; 2) the inability for generating traffic for future dates. The selected data source for this investigation has been selected explicitly to not have missing data, so traffic recordings are available for the NSE generative approach. Likewise, aggregation errors and other data anomalies have been cleaned. Otherwise, the NSE method would have output these corrupted data as synthetic samples, producing inaccurate predictions. The second concern might not be critical for some implementations, as the system is aimed at data generation. However, due to the nature of traffic data, synthetic samples can also serve as future traffic estimations, so the inability to be used for traffic prediction might be a hindrance for the NSE approach. Nevertheless, both issues emanate from the nature of the technique itself, thus being unbridgeable. At this juncture, we advocate for the use of the Cluster method. While still being simple to implement, delivered results are close to the NSE approach without the inconveniences of it.

The Cluster approach could provide enhanced performance with a dedicated set of grouping criteria. For the sake of simplicity and not delving into the particularities of the traffic behavior at the city of Madrid, we have presented an approach based on weekdays and holidays only. Still, how to group available training samples is the key element towards optimal performance. Daily traffic patterns can be grouped considering additional criteria such as weather or events (for instance, football matches). Another option is to perform an autonomous search of clusters with tools like DBSCAN or K-means \cite{khan2014dbscan,likas2003global}, which find traffic samples with a similar flow traffic pattern, without considering other criteria. A final innovative approach could be focusing on the search of atypical traffic samples sizes. Traffic flow follows a daily pattern, but this profile can also be fragmented into shorter segments (e.g., hourly segments). While early morning flow is often shared among all days, the flow spike observed during rush hours can present distinct shapes. In this way, partial daily segments could produce further clusters, providing more accurate predictions. In conclusion, the goal should be to produce clusters where all samples have a similar traffic distribution.

\begin{table}[ht]
\centering
    \begin{adjustbox}{max width=\columnwidth}
    \begin{threeparttable}
        \caption{Summary of sensed locations}
        \label{tab: detail}
        \begin{tabular}{rccc} 
         \multirow{1}{*}{}  &
         \multicolumn{1}{c}{\textbf{Secondary}} &
         \multicolumn{1}{c}{\textbf{Tertiary}}  &
         \multicolumn{1}{c}{\textbf{Residential}} 
         \\
         
         \cmidrule(lr){2-2}
         \cmidrule(lr){3-3}
         \cmidrule(lr){4-4}
         
         
         \textbf{\# sensed locations} & 23 & 14 & 18  \\
         \textbf{\# GAN as the best approach} & 3 & 0 & 6\\
         \textbf{\# Cluster as the best approach} & 4 & 3 & 8\\
         \textbf{\# NSE as the best approach} & 20 & 12 & 11\\
         \cmidrule(lr){1-4}
         \textbf{Mean nRMSE} & 0.50 & 0.53 & 0.75  \\
         
    \end{tabular}
        
  \end{threeparttable}
  \end{adjustbox}
\end{table}

The contents of Table \ref{tab: detail} are extracted from Table \ref{tab: generation}. It is intended to summarize certain statistics drawn from the road type of the target locations. Among considered road segments, road types are distributed as follows: 42\% secondary, 25\% tertiary, and 33\% residential. However, the mean performance error is not evenly distributed, where the synthetic data for residential roads is more different than real flow measurements. Coincidentally, whereas the NSE approach is the preferred option for secondary and tertiary roads, residential roads do not rely on NSE and also select the best generative approach the GAN and Cluster methods. From Figure \ref{fig:rmse} it can be seen that the mean flow of residential roads is usually below 200 vehicles. As the nRMSE is normalized by this last metric, it narrows the room for generation mistakes. This fact leads to the real bottleneck towards a leading performance: the selection of the sensed road segment. Of course, the same error can be committed for other road types, but after normalization, the nRMSE for residential roads is going to be higher. When an appropriate sensed road segment is selected, the NSE approach is most of the time the method that delivers the best performance. This behavior is motivated by the functioning of the generation method, where real traffic flow measurements serve as synthetic data. In contrast, the GAN and Cluster approaches output a smoothed traffic pattern. If the target road segments are analogous to the selected sensed location, the details present in the NSE output provide a high-quality prediction. On the opposite, after a poor choice of sensed road segment, a less disordered traffic pattern performs better.

In essence, the generative approach is not what determines the quality of the synthetic samples, but the selection of the road segment that provides the training data. 

\subsection*{Discussion and limitations}

Several conclusions can be drawn from this experimentation. To begin with, we have verified that the tailored selection of the segment providing traffic data for the estimation at the target location outperforms a na\"ive selection method based on geographical proximity. This stresses on the importance of topological and graph-based features to choose, in an informed fashion, which traffic data to use along the estimation process. This has been noted to be more decisive than the approach used to generate the traffic profile in the target location. As a matter of fact, results have revealed that a na\"ive estimation method based on assigning the traffic measurements collected in the other most similar location for the same date performs relatively better than other more sophisticated and less interpretable model-based approaches, further highlighting the significance of our proposed road feature embeddings for its competitive performance and its inherent algorithmic transparency.

Despite the higher trustworthiness derived from its inception from domain-specific knowledge and intuition, an evident limitation of considering only topological and graph-based features in the estimation of traffic profiles is that other exogenous factors are not explicitly considered along the process. Instead, aspects such as sociological habits and cultural traits are assumed to be embedded in the traffic data itself, as well as in the experience of traffic experts from which road features were formulated. This implicitness may not affect when estimating data \emph{inside} a city, or a neighborhood, but may conversely hinder the transferability of traffic profiles across different cities even if strongly similar road feature embeddings are discovered. In other words, estimating traffic profiles at one city using data collected by sensors of another city might not be straightforward due to the aforementioned factors. This last observation echoes our previous statement about the generalization of this study to account for other sources of information which, in the light of our experiments, seems to be crucial for an accurate shaping of the traffic profiles in the location of interest.

\section{Conclusions and future work}\label{sec:conclusions}

This work has elaborated on the problem of estimating road traffic data over a location of a road network without any deployed sensor nor prior collected traffic data in the location whatsoever. Under these circumstances, it becomes necessary to resort to other sources of information to estimate the traffic profiles occurring at the target location. For this purpose, this manuscript has proposed to exploit topological and graph-based information around the location of interest to find other \emph{similar} road segments. The overarching idea is that traffic data collected over those similar segments can be used to estimate the traffic in the target location. To compute the similarity between any two road segments, we characterize each location in the form of road feature embeddings, built upon a set of topological and graph-based features. The definitions of these features rely on intuition and domain-specific knowledge about the traffic dynamics in the location of interest, which ultimately adds to the overall trustworthiness of traffic estimations by decision makers for which they are produced. Such road embeddings are used together with a measure of distance to compute the similarity between the segments at hand, so that the location closest to the target can be identified and used for traffic estimation. It is important to note that the challenging assumptions from which our work departs (unavailability of any other source of information beyond the road network) do not hinder the generalization of our developments to other scenarios with further information available for the traffic estimation process. 
Nevertheless, features within the road feature embedding should be designed according to the knowledge of traffic managers regarding the target area, in a similar process to the one in Madrid (Spain) showcased in this study. We have herein focused on the traffic map of a major city, but urban areas in the countryside connected by interurban networks might require further investigations to yield new topological and graph-based features tailored for such particular scenarios.

Besides the limitation just exposed, other research lines rooted on this work are projected for the near future. We plan to investigate new and more insightful road embedding features, potentially using meta-learning methods to infer them automatically from data (e.g., symbolic regression, evolutionary programming with graph-based primitives). In partial connection with this, we also foresee that a generalized distance metric – for instance, a weighted Euclidean distance – could be evolved via an evolutionary wrapper towards optimally tuning the importance of every feature of the embedding in the value of the distance. Finally, other sources of information that can be retrieved from geographical information systems can be valuable inputs to be considered in our road feature embeddings, without drifting away from the assumed starting point of this work. As such, hospitals and police stations can regulate the average speed of the surrounding roads under the legal limit, whereas entertainment venues surrounded by stores and restaurants often have a high road occupation percentage on holidays and weekends. The \emph{Santiago Bernabeu} stadium is located inside the area of the case study. Football matches affect the traffic of the surrounding roads, where the intensity of its influence is inversely related to the distance to the stadium \cite{olabarrieta2020effect}. The so-called \emph{Points of Interest} (POIs) comprise any kind of business, service center or important area that might influence the traffic behavior. Consequently, future work will embrace such POI-based information (when available) to produce better road feature embeddings that leverage even further the existence of expert knowledge about the factors affecting the traffic at the target location. We envision that the precision of the traffic estimation process can be improved further by extending the embedding defined herein with graph features that depend on the relative location of these POIs
with respect to the location for which traffic data is to be estimated. Therefore, much of our future effort will be conducted towards this promising research direction.

\section*{Acknowledgments}

The authors would like to thank the Basque Government for its funding support through the EMAITEK and ELKARTEK programs (3KIA project, KK-2020/00049). Eric L. Manibardo receives funding support from the Basque Government through its BIKAINTEK PhD support program (grant no. 48AFW22019-00002). Javier Del Ser also thanks the same institution for the funding support received through the consolidated research group MATHMODE (ref. IT1456-22).

\bibliographystyle{elsarticle-num} 
\bibliography{referencias_V2}

\begin{thebibliography}{10}
\expandafter\ifx\csname url\endcsname\relax
  \def\url#1{\texttt{#1}}\fi
\expandafter\ifx\csname urlprefix\endcsname\relax\def\urlprefix{URL }\fi
\expandafter\ifx\csname href\endcsname\relax
  \def\href#1#2{#2} \def\path#1{#1}\fi

\bibitem{lana2018road}
I.~La{\~n}a, J.~Del~Ser, M.~Velez, E.~I. Vlahogianni, Road traffic forecasting:
  Recent advances and new challenges, IEEE Intelligent Transportation Systems
  Magazine 10~(2) (2018) 93--109.

\bibitem{manibardo2020transfer}
E.~L. Manibardo, I.~La{\~n}a, J.~Del~Ser, Transfer learning and online learning
  for traffic forecasting under different data availability conditions:
  Alternatives and pitfalls, in: IEEE International Conference on Intelligent
  Transportation Systems (ITSC), IEEE, 2020, pp. 1--6.

\bibitem{manibardo2021deep}
E.~L. Manibardo, I.~La{\~n}a, J.~Del~Ser, Deep learning for road traffic
  forecasting: Does it make a difference?, IEEE Transactions on Intelligent
  Transportation Systems (2021).

\bibitem{pipino2002data}
L.~L. Pipino, Y.~W. Lee, R.~Y. Wang, Data quality assessment, Communications of
  the ACM 45~(4) (2002) 211--218.

\bibitem{PeMSddbb}
{Caltrans, Performance Measurement System}, \url{http://pems.dot.ca.gov},
  accessed: 2020-11-06.

\bibitem{liao2018deep}
B.~Liao, J.~Zhang, C.~Wu, D.~McIlwraith, T.~Chen, S.~Yang, Y.~Guo, F.~Wu, Deep
  sequence learning with auxiliary information for traffic prediction, in: ACM
  SIGKDD International Conference on Knowledge Discovery \& Data Mining, 2018,
  pp. 537--546.

\bibitem{jagadish2014big}
H.~V. Jagadish, J.~Gehrke, A.~Labrinidis, Y.~Papakonstantinou, J.~M. Patel,
  R.~Ramakrishnan, C.~Shahabi, Big data and its technical challenges,
  Communications of the ACM 57~(7) (2014) 86--94.

\bibitem{lana2021data}
I.~La{\~n}a, J.~J. Sanchez-Medina, E.~I. Vlahogianni, J.~Del~Ser, From data to
  actions in {I}ntelligent {T}ransportation {S}ystems: a prescription of
  functional requirements for model actionability, Sensors 21~(4) (2021) 1121.

\bibitem{lana2021soft}
I.~La{\~n}a, I.~Oregi, J.~Del~Ser, Soft sensing methods for the generation of
  plausible traffic data in sensor-less locations, in: IEEE International
  Intelligent Transportation Systems Conference (ITSC), IEEE, 2021, pp.
  3183--3189.

\bibitem{vlahogianni2014short}
E.~I. Vlahogianni, M.~G. Karlaftis, J.~C. Golias, Short-term traffic
  forecasting: Where we are and where we’re going, Transportation Research
  Part C: Emerging Technologies 43 (2014) 3--19.

\bibitem{brinkhoff2003generating}
T.~Brinkhoff, Generating traffic data, IEEE Computer Society Technical
  Committee on Data Engineering 26~(2) (2003) 19--25.

\bibitem{krajzewicz2012recent}
D.~Krajzewicz, J.~Erdmann, M.~Behrisch, L.~Bieker, Recent development and
  applications of {SUMO}-simulation of urban mobility, International Journal on
  Advances in Systems and Measurements 5~(3\&4) (2012).

\bibitem{fellendorf2010microscopic}
M.~Fellendorf, P.~Vortisch, Microscopic traffic flow simulator {VISSIM}, in:
  Fundamentals of traffic simulation, Springer, 2010, pp. 63--93.

\bibitem{owen2000traffic}
L.~E. Owen, Y.~Zhang, L.~Rao, G.~McHale, Traffic flow simulation using
  {CORSIM}, in: 2000 Winter Simulation Conference Proceedings, Vol.~2, IEEE,
  2000, pp. 1143--1147.

\bibitem{w2016multi}
K.~W~Axhausen, A.~Horni, K.~Nagel, The multi-agent transport simulation MATSim,
  Ubiquity Press, 2016.

\bibitem{lopez2018microscopic}
P.~A. Lopez, M.~Behrisch, L.~Bieker-Walz, J.~Erdmann, Y.-P. Fl{\"o}tter{\"o}d,
  R.~Hilbrich, L.~L{\"u}cken, J.~Rummel, P.~Wagner, E.~Wie{\ss}ner, Microscopic
  traffic simulation using {SUMO}, in: 2018 21st International Conference on
  Intelligent Transportation Systems (ITSC), IEEE, 2018, pp. 2575--2582.

\bibitem{reilly2009bangalore}
M.~K. Reilly, M.~P. O’Mara, K.~C. Seto, From {B}angalore to the {B}ay area:
  Comparing transportation and activity accessibility as drivers of urban
  growth, Landscape and Urban Planning 92~(1) (2009) 24--33.

\bibitem{wen2017understanding}
T.-H. Wen, P.-C. Lai, et~al., Understanding the topological characteristics and
  flow complexity of urban traffic congestion, Physica A: Statistical Mechanics
  and its Applications 473 (2017) 166--177.

\bibitem{wang2018analyzing}
S.~Wang, D.~Yu, X.~Ma, X.~Xing, Analyzing urban traffic demand distribution and
  the correlation between traffic flow and the built environment based on
  detector data and {POI}s, European Transport Research Review 10~(2) (2018)
  1--17.

\bibitem{geroliminis2008existence}
N.~Geroliminis, C.~F. Daganzo, Existence of urban-scale macroscopic fundamental
  diagrams: Some experimental findings, Transportation Research Part B:
  Methodological 42~(9) (2008) 759--770.

\bibitem{ambuhl2021disentangling}
L.~Amb{\"u}hl, A.~Loder, L.~Leclercq, M.~Menendez, Disentangling the city
  traffic rhythms: A longitudinal analysis of mfd patterns over a year,
  Transportation Research Part C: Emerging Technologies 126 (2021) 103065.

\bibitem{laval2015stochastic}
J.~A. Laval, F.~Castrill{\'o}n, Stochastic approximations for the macroscopic
  fundamental diagram of urban networks, Transportation Research Procedia 7
  (2015) 615--630.

\bibitem{wong2021network}
W.~Wong, S.~Wong, H.~X. Liu, Network topological effects on the macroscopic
  fundamental diagram, Transportmetrica B: Transport Dynamics 9~(1) (2021)
  376--398.

\bibitem{sirmatel2021stabilization}
I.~I. Sirmatel, N.~Geroliminis, Stabilization of city-scale road traffic
  networks via macroscopic fundamental diagram-based model predictive perimeter
  control, Control Engineering Practice 109 (2021) 104750.

\bibitem{lecun2015deep}
Y.~LeCun, Y.~Bengio, G.~Hinton, Deep learning, Nature 521~(7553) (2015)
  436--444.

\bibitem{landolt2021taxonomy}
S.~Landolt, T.~Wambsganss, M.~S{\"o}llner, A taxonomy for deep learning in
  natural language processing, Hawaii International Conference on System
  Sciences (ICSS), 2021.

\bibitem{hassaballah2020deep}
M.~Hassaballah, A.~I. Awad, Deep learning in computer vision: principles and
  applications, CRC Press, 2020.

\bibitem{darmochwal1991euclidean}
A.~Darmochwa{\l}, The {E}uclidean space, Formalized Mathematics 2~(4) (1991)
  599--603.

\bibitem{vaswani2017attention}
A.~Vaswani, N.~Shazeer, N.~Parmar, J.~Uszkoreit, L.~Jones, A.~N. Gomez,
  {\L}.~Kaiser, I.~Polosukhin, Attention is all you need, Advances in Neural
  Information Processing Systems 30 (2017).

\bibitem{arrieta2020explainable}
A.~B. Arrieta, N.~D{\'\i}az-Rodr{\'\i}guez, J.~Del~Ser, A.~Bennetot, S.~Tabik,
  A.~Barbado, S.~Garc{\'\i}a, S.~Gil-L{\'o}pez, D.~Molina, R.~Benjamins,
  et~al., Explainable artificial intelligence ({XAI}): Concepts, taxonomies,
  opportunities and challenges toward responsible {AI}, Information Fusion 58
  (2020) 82--115.

\bibitem{bronstein2017geometric}
M.~M. Bronstein, J.~Bruna, Y.~LeCun, A.~Szlam, P.~Vandergheynst, Geometric deep
  learning: going beyond {E}uclidean data, IEEE Signal Processing Magazine
  34~(4) (2017) 18--42.

\bibitem{harary2018graph}
F.~Harary, Graph theory (2018).

\bibitem{munkres2018analysis}
J.~R. Munkres, Analysis on manifolds, CRC Press, 2018.

\bibitem{wu2020comprehensive}
Z.~Wu, S.~Pan, F.~Chen, G.~Long, C.~Zhang, S.~Y. Philip, A comprehensive survey
  on graph neural networks, IEEE Transactions on Neural Networks and Learning
  Systems 32~(1) (2020) 4--24.

\bibitem{jiang2021graph}
W.~Jiang, J.~Luo, Graph neural network for traffic forecasting: A survey, arXiv
  preprint arXiv:2101.11174 (2021).

\bibitem{Madridddbb}
{Madrid Open Data Portal}, \url{http://datos.madrid.es}, accessed: 2020-11-06.

\bibitem{boeing2017osmnx}
G.~Boeing, {OSMnx}: New methods for acquiring, constructing, analyzing, and
  visualizing complex street networks, Computers, Environment and Urban Systems
  65 (2017) 126--139.

\bibitem{OpenStreetMap}
{OpenStreetMap contributors}, {Planet dump retrieved from
  https://planet.osm.org }, \url{ https://www.openstreetmap.org } (2017).

\bibitem{sandfelder2021ego}
D.~Sandfelder, P.~Vijayan, W.~L. Hamilton, Ego-{GNNs}: Exploiting ego
  structures in graph neural networks, in: IEEE International Conference on
  Acoustics, Speech and Signal Processing (ICASSP), IEEE, 2021, pp. 8523--8527.

\bibitem{hagberg2008exploring}
A.~Hagberg, P.~Swart, D.~S~Chult, Exploring network structure, dynamics, and
  function using {NetworkX}, Tech. rep., Los Alamos National Lab.(LANL), Los
  Alamos, NM (United States) (2008).

\bibitem{wasserman1994social}
S.~Wasserman, K.~Faust, et~al., Social network analysis: Methods and
  applications (1994).

\bibitem{holme2003congestion}
P.~Holme, Congestion and centrality in traffic flow on complex networks,
  Advances in Complex Systems 6~(02) (2003) 163--176.

\bibitem{valente2008correlated}
T.~W. Valente, K.~Coronges, C.~Lakon, E.~Costenbader, How correlated are
  network centrality measures?, Connections (Toronto, Ont.) 28~(1) (2008) 16.

\bibitem{brandes2001faster}
U.~Brandes, A faster algorithm for betweenness centrality, Journal of
  Mathematical Sociology 25~(2) (2001) 163--177.

\bibitem{braess2005paradox}
D.~Braess, A.~Nagurney, T.~Wakolbinger, On a paradox of traffic planning,
  Transportation Science 39~(4) (2005) 446--450.

\bibitem{tabak2014geometry}
J.~Tabak, Geometry: the language of space and form, Infobase Publishing, 2014.

\bibitem{lana2016understanding}
I.~Lana, J.~Del~Ser, I.~I. Olabarrieta, Understanding daily mobility patterns
  in urban road networks using traffic flow analytics, in: NOMS 2016-2016
  IEEE/IFIP Network Operations and Management Symposium, IEEE, 2016, pp.
  1157--1162.

\bibitem{lana2019adaptive}
I.~La{\~n}a, J.~L. Lobo, E.~Capecci, J.~Del~Ser, N.~Kasabov, Adaptive long-term
  traffic state estimation with evolving spiking neural networks,
  Transportation Research Part C: Emerging Technologies 101 (2019) 126--144.

\bibitem{mirza2014conditional}
M.~Mirza, S.~Osindero, Conditional generative adversarial nets, arXiv preprint
  arXiv:1411.1784 (2014).

\bibitem{manibardo2021change}
E.~L. Manibardo, I.~La{\~n}a, J.~Del~Ser, Change detection and adaptation
  strategies for long-term estimation of pedestrian flows, in: IEEE
  International Intelligent Transportation Systems Conference (ITSC), IEEE,
  2021, pp. 1867--1874.

\bibitem{friedman1940comparison}
M.~Friedman, A comparison of alternative tests of significance for the problem
  of m rankings, The Annals of Mathematical Statistics 11~(1) (1940) 86--92.

\bibitem{nemenyi1963distribution}
P.~B. Nemenyi, Distribution-free multiple comparisons., Princeton University,
  1963.

\bibitem{khan2014dbscan}
K.~Khan, S.~U. Rehman, K.~Aziz, S.~Fong, S.~Sarasvady, {DBSCAN}: Past, present
  and future, in: International Conference on the Applications of Digital
  Information and Web Technologies (ICADIWT), IEEE, 2014, pp. 232--238.

\bibitem{likas2003global}
A.~Likas, N.~Vlassis, J.~J. Verbeek, The global k-means clustering algorithm,
  Pattern recognition 36~(2) (2003) 451--461.

\bibitem{olabarrieta2020effect}
I.~I. Olabarrieta, I.~La{\~n}a, Effect of soccer games on traffic, study case:
  Madrid, in: 2020 IEEE 23rd International Conference on Intelligent
  Transportation Systems (ITSC), IEEE, 2020, pp. 1--5.

\end{thebibliography}

\begin{IEEEbiography}[{\includegraphics[width=1in,clip,keepaspectratio]{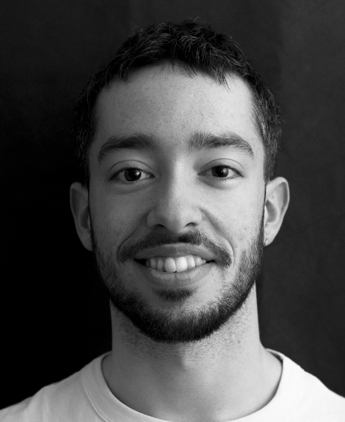}}]{Eric L. Manibardo} received his B.Sc. degree in Telecommunication Engineering in 2017, and M.Sc. degree also in Telecommunications Engineering in 2019 from the University of the Basque Country, Spain. He is currently a junior researcher at TECNALIA (Spain), pursuing his PhD in Artificial Intelligence. His research interest combine machine learning and signal processing within the context of Intelligent Transportation Systems (ITS), with an emphasis on traffic forecasting. 
\end{IEEEbiography}

\begin{IEEEbiography}[{\includegraphics[width=1in,clip,keepaspectratio]{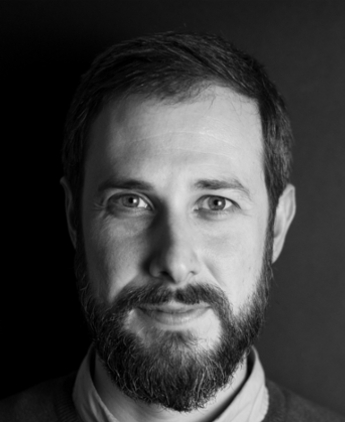}}]{Ibai La\~na} received his B.Sc. degree in Computer Engineering from Deusto University, Spain, in 2006, the M.Sc. degree in Advanced Artificial Intelligence from UNED, Spain, in 2014, and the PhD in Artificial Intelligence from the University of the Basque Country in 2018. He is currently a senior researcher at TECNALIA (Spain).  His research interests fall within the intersection of Intelligent Transportation Systems (ITS), machine learning, traffic data analysis and data science. He has dealt with urban traffic forecasting problems, where he has applied machine learning models and evolutionary algorithms to obtain longer term and more accurate predictions. He also has interest in other traffic related challenges, such as origin-destination matrix estimation or point of interest and trajectory detection.
\end{IEEEbiography}

\begin{IEEEbiography}[{\includegraphics[width=1in,clip,keepaspectratio]{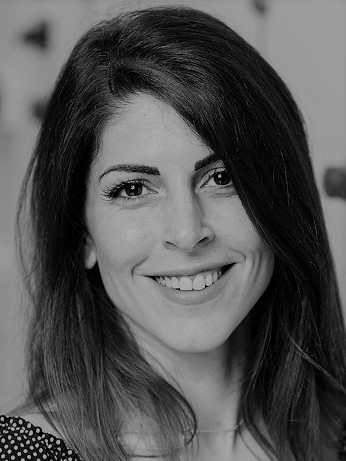}}]{Esther Villar-Rodriguez} holds a PhD (Cum Laude) in Information and Communication Technologies (2015) from the University of Alcala (Spain), a BSc in Computer Science (2010) by the University of Deusto, and a M.Sc. (2012) in Computer Languages and Systems by UNED (National University of Distance Education, Spain). She is currently a Principal Researcher in Artificial Intelligence at TECNALIA (Spain), specialized in Shallow Learning and Deep Machine Learning models, distributed privacy-aware learning strategies (including Federated Learning), Reinforcement Learning and outlier detection, among others. She has authored several contributions at conferences and articles in journals related to these research areas, with a focus on their applicability to practical problems (prediction and optimization of management and industrial processes, resource planning, scheduling, energy efficiency, and other assorted applications alike). 
\end{IEEEbiography}
	
\begin{IEEEbiography}[{\includegraphics[width=1in,clip,keepaspectratio]{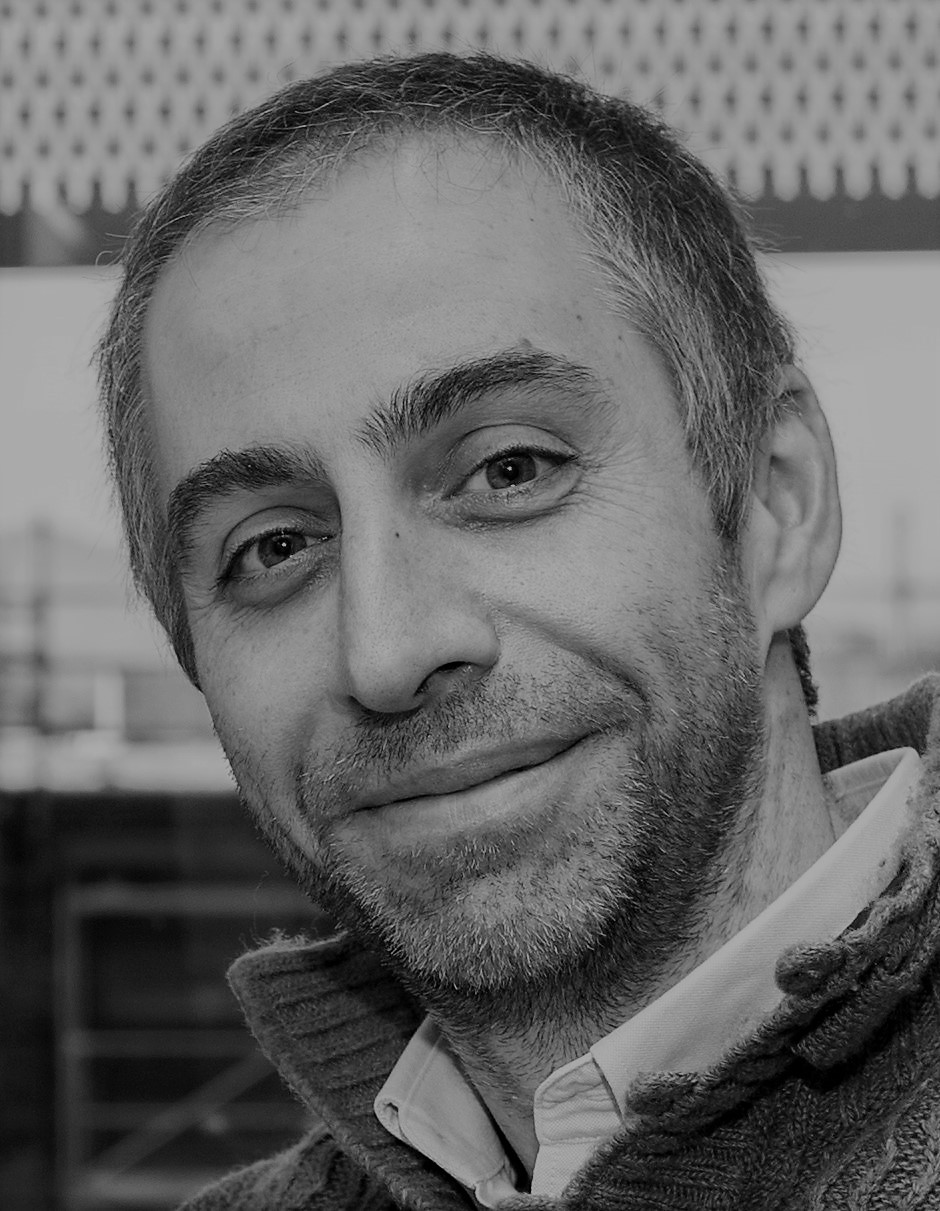}}]{Javier Del Ser} (M'07-SM'12) received his first PhD in Telecommunication Engineering (Cum Laude) from the University of Navarra, Spain, in 2006, and a second PhD in Computational Intelligence (Summa Cum Laude, Extraordinary PhD Prize) from the University of Alcala, Spain, in 2013. He is currently a Research Professor in data analytics and optimization at TECNALIA (Spain) and an adjunct professor at the University of the Basque Country (UPV/EHU). He has published more than 400 journal articles, book chapters and conference contributions, co-supervised 11 Ph.D. theses, edited 6 books and co-invented 9 patents in the broad topics of Artificial Intelligence, Data Science and Optimization. He is associate editor of several journals related to areas of Artificial Intelligence, including Swarm and Evolutionary Computation, Information Fusion, Cognitive Computation and IEEE Transactions on ITS.
\end{IEEEbiography}
\vfill

\end{document}